\documentclass[11pt,a4paper]{article}
\usepackage{times,latexsym}
\usepackage{url}
\usepackage[T1]{fontenc}
\usepackage{booktabs}
\usepackage{graphicx}
\usepackage{multirow}

\usepackage[acceptedWithA]{tacl2018v2}

\usepackage{xspace,mfirstuc,tabulary}

\newif\iftaclinstructions
\taclinstructionsfalse 
\iftaclinstructions

\renewcommand{\anonsubtext}{(No author info supplied here, for consistency with
TACL-submission anonymization requirements)}
\newcommand{\instr}
\fi

\iftaclpubformat 

\else

\fi

\title{Experts, Errors, and Context:\\A Large-Scale Study of Human Evaluation for Machine Translation}

\author{Markus Freitag, George Foster, David Grangier, Viresh Ratnakar, Qijun Tan, Wolfgang Macherey \\
  Google Research \\
\texttt{\{freitag, fosterg, grangier, vratnakar, qijuntan, wmach\}@google.com} \\}

\date{}

\begin{document}
\maketitle
\begin{abstract}

Human evaluation of modern high-quality machine translation systems is a difficult problem, and there is increasing evidence that inadequate evaluation procedures can lead to erroneous conclusions. While there has been considerable research on human evaluation, the field still lacks a commonly-accepted standard procedure. As a step toward this goal, we propose an evaluation methodology grounded in explicit error analysis, based on the Multidimensional Quality Metrics (MQM) framework.  We carry out the largest MQM research study to date, scoring the outputs of top systems from the WMT 2020 shared task in two language pairs using annotations provided by professional translators with access to full document context. We analyze the resulting data extensively, finding among other results a substantially different ranking of evaluated systems from the one established by the WMT crowd workers, exhibiting a clear preference for human over machine output. Surprisingly, we also find that automatic metrics based on pre-trained embeddings can outperform human crowd workers. We make our corpus publicly available for further research.

\end{abstract}

\section{Introduction}

Like many natural language generation tasks, machine translation (MT) is difficult to evaluate because the set of correct answers for each input is large and usually unknown. This limits the accuracy of automatic metrics, and necessitates costly human evaluation to provide a reliable gold standard for measuring MT quality and progress. Yet even human evaluation is problematic. For instance, we often wish to decide which of two translations is better, and by how much, but what should this take into account? If one translation sounds somewhat more natural than another, but contains a slight inaccuracy, what is the best way to quantify this? To what extent will different raters agree on their assessments?

The complexities of evaluating translations---both machine and human---have been extensively studied, and there are many recommended best practices. However, due to expedience, human evaluation of MT is frequently carried out on isolated sentences by inexperienced raters with the aim of assigning a single score or ranking. When MT quality is poor, this can provide a useful signal; but as quality improves, there is a risk that the signal will become lost in rater noise or bias. Recent papers have argued that poor human evaluation practices have led to misleading results, including erroneous claims that MT has achieved human parity \cite{toral2020reassessing, laubli2018has}.

This paper aims to contribute to the evolution of standard practices for human evaluation of high-quality MT. Our key insight is that any scoring or ranking of translations is implicitly based on an identification of errors and other imperfections. Making such an identification explicit by enumerating errors provides a ``platinum standard’’ from which various gold-standard scorings can be derived, depending on the importance placed on different categories of errors for different downstream tasks. This is not a new insight: it is the conceptual basis for the Multidimensional Quality Metrics (MQM) framework developed in the EU QTLaunchPad and QT21 projects \href{www.qt21.eu}{(www.qt21.eu)}, which we endorse and adopt for our experiments.

MQM is a generic framework that provides a hierarchy of translation errors which can be tailored to specific applications. We identified a hierarchy appropriate for broad-coverage MT, and annotated outputs from 10 top-performing "systems" (including human references) for both the English$\to$German (EnDe) and Chinese$\to$English (ZhEn) language directions in the WMT 2020 news translation task~\cite{barrault-etal-2020-findings}, using professional translators with access to full document context. For comparison purposes, we also collected scalar ratings on a 7-point scale from both professionals and crowd workers.

We analyze the resulting data along many different dimensions: comparing the system rankings resulting from different rating methods, including the original WMT scores; characterizing the error patterns of modern neural MT systems, including profiles of difficulty across documents, and comparing them to human translations (HT);  measuring MQM inter-annotator agreement; and re-evaluating the performance of automatic metrics submitted to the WMT 2020 metrics task. Our most striking finding is that MQM ratings sharply revise the original WMT ranking of translations, exhibiting a clear preference for HT over MT, and promoting some low-ranked MT systems to much higher positions. This in turn changes the conclusions about the relative performance of different automatic metrics; interestingly, we find that most metrics correlate better with MQM rankings than WMT human scores do. We hope these results will underscore and help publicize the need for more careful human evaluation, particularly in shared tasks intended to assess MT or metric performance. We release our corpus to encourage further research.\footnote{\url{https://github.com/google/wmt-mqm-human-evaluation}}
Our main contributions are:
\begin{itemize}
\item A proposal for a standard MQM scoring scheme appropriate for broad-coverage MT.
\item Release of a large-scale MQM corpus with annotations for over 100k HT and high-quality-MT segments in two language pairs (EnDe and ZhEn) from WMT 2020. This is by far the largest study of human evaluation results released to the public.
\item Re-evaluation of the performance of MT systems and automatic metrics on our corpus, showing clear distinctions between HT and MT based on MQM ratings, adding to the evidence against claims of human parity.
\item Demonstration that crowd-worker evaluation has low correlation with our MQM-based evaluation, calling into question conclusions drawn on the basis of previous crowd-sourced evaluations.
\item Demonstration that automatic metrics based on pre-trained embeddings can outperform human crowd workers.
\item Characterization of current error types in HT and MT, identifying specific MT weaknesses.
\item Recommendations for the number of ratings needed to establish a reliable human benchmark, and for the most efficient way of distributing them across documents.
\end{itemize}

\section{Related Work}

One of the earliest formal mentions of human evaluation for MT occurs in the ALPAC report \shortcite{national1966language}, which defines an evaluation methodology based on “intelligibility” (comprehensibility) and “fidelity” (adequacy). The ARPA MT Initiative \cite{white1994arpa} defines an overall quality score based on “adequacy”, “fluency” and “comprehension”.
In 2006, the first WMT evaluation campaign \cite{koehn2006manual} used adequacy and fluency ratings on a 5 point scale acquired from participants as their main metric.
\newcite{vilar2007human} proposed a ranking-based evaluation approach which became the official metric at WMT from 2008 until 2016 \cite{callison2008proceedings}. The ratings were still acquired from the participants of the evaluation campaign.
\newcite{graham2013continuous} compared human assessor consistency levels for judgments collected on a five-point
interval-level scale to those collected on a 1-100 continuous scale, using machine translation fluency as
a test case. They claim that the use of a continuous scale eliminates individual judge preferences, resulting in higher levels of inter-annotator consistency.
\newcite{bojar2016wmt} came to the conclusion that fluency evaluation is highly correlated to adequacy evaluation. As a consequence of the latter two papers, continuous direct assessment focusing on adequacy has been the official WMT metric since 2017 \cite{ondrej2017findings}. Due to budget constraints, WMT understandably conducts its human evaluation with researchers and/or crowd-workers.

\newcite{avramidis-etal-2012-involving} used professional translators to rate MT output on three different tasks: ranking, error classification and post-editing. \newcite{castilho2017comparative} found that crowd workers lack knowledge of translation and, compared to professional translators, tend to be more accepting of (subtle) translation errors. \newcite{graham2017can} showed that crowd-worker evaluation has to be filtered to avoid contamination of results through the inclusion of false assessments.
The quality of ratings acquired by either researchers or crowd workers has further been questioned by \cite{Toral18,laubli2020recommendations}, who demonstrated that professional translators can discriminate between human and machine translations where crowd-workers were not able to do so.
\newcite{mathur-etal-2020-results} re-evaluated a subset of WMT submissions with professional translators and showed that the resulting rankings changed and were better aligned with automatic scores. \newcite{fischer-laubli-2020-whats} found that the number of segments with wrong terminology, omissions, and typographical problems for MT output is similar to HT.
\newcite{fomicheva2017role,bentivogli2018machine} raised the concern that reference-based human evaluation might penalise correct translations that diverge too much from the reference. The literature mostly agrees that source-based rather than reference-based evaluation should be conducted \cite{laubli2020recommendations}.
The impact of translationese~\cite{Koppel:2011:TD:2002472.2002636} on human evaluation of MT has recently received attention~\citep{Toral18,zhang2019effect,Freitag19,graham2020translationese}. These papers show that the nature of source sentences is important and that only natural source sentences should be used for human evaluation. 

As alternatives to adequacy and fluency, \newcite{scarton2016reading} presented reading comprehension for MT quality evaluation. \newcite{forcada2018exploring} proposed gap-filling, where certain words are removed from reference translations and readers are asked to fill the gaps left using the machine-translated text as a hint. \newcite{popovic2020informative} proposed a new method for manual evaluation based on marking actual issues in the translated text. Instead of assigning a score, annotators are asked to just label problematic parts of the translations.

The Multidimensional Quality Metrics (MQM) framework was developed in the EU QTLaunchPad and QT21 projects (2012--2018) \href{www.qt21.eu}{(www.qt21.eu)} to address the shortcomings of previous quality evaluation methods \cite{lommel2014multidimensional}.
MQM provides a generic methodology for assessing translation quality that can be adapted to a wide range of evaluation needs. 
\newcite{klubivcka2018quantitative} designed an MQM-compliant error taxonomy tailored to the relevant linguistic phenomena of Slavic languages to run a case study for 3 MT systems for English$\to$Croatian. More recently, \newcite{rei-etal-2020-comet} used MQM labels to fine-tune COMET for automatic evaluation.

\section{Human Evaluation Methodologies}

We compared three human evaluation techniques: the WMT 2020 baseline; ratings on a 7-point Likert-type scale which we refer to as a Scalar Quality Metric (SQM); and evaluations under the MQM framework. We describe these methodologies in the following three sections, deferring concrete experimental details about annotators and data to the subsequent section.

\subsection{WMT}
As part of the WMT evaluation campaign~\cite{barrault-etal-2020-findings}, WMT runs human evaluation of the primary submissions for each language pair. The organizers collect segment-level ratings with document context (SR+DC) on a 0-100 scale using either source-based evaluation with a mix of researchers/translators (for translations out of English) or reference-based evaluation with crowd-workers (for translations into English). In addition, WMT conducts rater quality controls to remove ratings from raters that are not trustworthy. In general, for each system, only a subset of documents receive ratings, with the rated subset differing across systems. The organizers provide two different segment-level scores, averaged across one or more raters: (a) the raw score; and (b) a z-score which is standardized for each annotator. Document- and system-level scores are averages over segment-level scores. For more details, we refer the reader to the WMT findings papers.

\subsection{SQM}
Similar to the WMT setting, the Scalar Quality Metric (SQM) evaluation collects segment-level scalar ratings with document context. 
Different from the 0-100 assessment of translation quality used in WMT,
SQM uses a 0-6 scale for
translation quality assessment, with the quality levels described as follows:

6: Perfect Meaning and Grammar:
The meaning of the translation is completely consistent with the source and the surrounding context (if applicable). The grammar is also correct.

4: Most Meaning Preserved and Few Grammar Mistakes:
The translation retains most of the meaning of the source. It may have some grammar mistakes or minor contextual inconsistencies.

2: Some Meaning Preserved:
The translation preserves some of the meaning of the source but misses significant parts. The narrative is hard to follow due to fundamental errors. Grammar may be poor.

0: Nonsense/No meaning preserved:
Nearly all information is lost between the translation and source. Grammar is irrelevant.

This evaluation presents each source segment and translated segment from a document in a table row, asking the rater to pick a rating from 0 through 6 (including the intermediate levels 1, 3, and 5). The rater can scroll up or down to see all the other source/translation segments from the document. 
Our SQM experiments used the 0-6 rating scale described above, instead of the wider, continuous scale recommended by \cite{graham2013continuous}, as this scale has been an established part of our existing MT evaluation ecosystem. It is possible that system rankings may be slightly sensitive to this nuance, but less so with raters who are translators rather than crowd workers, we believe. 

\subsection{MQM}

To adapt the generic MQM framework for our context, we followed the official guidelines for scientific research \href{http://qt21.eu/downloads/MQM-usage-guidelines.pdf}{(MQM-usage-guidelines.pdf)}. For space reasons we give only the salient features of our MQM customization here, referring the reader to appendix~\ref{sec:mqm-summary} for a summary of MQM, and to appendix~\ref{sec:mqm-details} for full details of our framework.

Our annotators were instructed to identify all errors within each segment in a document, paying particular attention to document context; see Table~\ref{tab:mqm-guidelines} for complete annotator guidelines. Each error was highlighted in the text, and labeled with an error category from Table~\ref{tab:mqm-hierarchy} and a severity from Table~\ref{tab:mqm-severity}. To temper the effect of long segments, we imposed a maximum of five errors per segment, instructing raters to choose the five most severe errors for segments containing more errors.

Our error hierarchy includes the standard top-level categories {\em Accuracy}, {\em Fluency}, {\em Terminology}, {\em Style}, and {\em Locale}, each with a specific set of  sub-categories. After an initial pilot run, we introduced a special {\em Non-translation} error that can be used to tag an entire segment which is too badly garbled to permit reliable identification of individual errors.

Error severities are assigned independent of category, and consist of {\em Major}, {\em Minor}, and {\em Neutral} levels, corresponding respectively to actual translation or grammatical errors, smaller imperfections, and purely subjective opinions about the translation. Many MQM schemes include an additional {\em Critical} severity which is worse than Major, but we dropped this because its definition is often context-specific. We felt that for broad coverage MT, the distinction between Major and Critical was likely to be highly subjective, while Major errors (true errors) would be easier to distinguish from Minor ones (imperfections).

Since we are ultimately interested in scoring segments, we require a weighting on error types. We fixed the weight on Minor errors at 1, and explored a range of Major weights from 1 to 10 (the Major weight recommended in the MQM standard). For each weight combination we examined the stability of system ranking using a resampling technique. We found that a Major weight of 5 gave the best balance between stability and ability to discriminate among systems.

These weights apply to all error categories with two exceptions. We assigned a weight of 0.1 to Minor Fluency/Punctuation errors to reflect their mostly non-linguistic nature. Decisions like the style of quotation mark to use or the spacing around punctuation affect the appearance of a text but do not change its meaning. Unlike other kinds of Minor errors, these are easy to correct algorithmically, so we assign a low weight to ensure that their main role is to distinguish between systems that are equivalent in other respects. Major Fluency/Punctuation errors, which render a text ungrammatical or change its meaning (eg, eliding the comma in “Let’s eat, grandma”), have standard weighting.
The second exception is the singleton Non-translation category, with a weight of 25, equivalent to five Major errors.

Table~\ref{tab:mqm-scoring} summarizes our weighting scheme, in which segment-level scores can range from 0 (perfect) to 25 (worst). The final segment-level score is an average over scores from all annotators.

\begin{table}[ht]\centering
\scalebox{0.80}{
\begin{tabular}{l|l|l}\toprule
Severity & Category & Weight \\
\midrule
Major & Non-translation & 25 \\
      & all others & 5 \\
\midrule
Minor & Fluency/Punctuation & 0.1 \\
      & all others          & 1 \\
\midrule
Neutral & all & 0 \\
\bottomrule
\end{tabular}
}
\caption{MQM error weighting.}
\label{tab:mqm-scoring}
\vspace{-1em}
\end{table}

\subsection{Experimental Setup}

We re-annotated the WMT 2020 English$\to$German and Chinese$\to$English test sets, comprising 1418 segments (130 documents) and 2000 segments (155 documents) respectively. For each set we chose 10 "systems" for annotation, including the three reference translations available for English$\to$German and the two references available for Chinese$\to$English. The MT outputs included all top-performing systems according to the WMT human evaluation, augmented with systems we selected to increase diversity. Tables~\ref{tab:ende_abs} and \ref{tab:chen_abs} list all evaluated systems.

Table~\ref{tab:annotated_dataset} summarizes rating information for the WMT evaluation and for the additional evaluations we conducted: SQM with crowd workers (cSQM), SQM with professional translators (pSQM), and MQM. We used disjoint professional translator pools for pSQM and MQM in order to avoid bias. All members of our rater pools were native speakers of the target language. Note that the average number of ratings per segment is less than 1 for the WMT evaluations because not all ratings survived the quality control.

\begin{table}[!htb]\centering
{\setlength{\tabcolsep}{.3em}
\begin{tabular}{l|rrr}\toprule
& ratings / seg & rater pool & raters \\
\midrule
WMT EnDe & 0.47 & res./trans. & 100 \\ 
WMT ZhEn & 0.86 & crowd & 115 \\ 
\midrule
cSQM EnDe & 1 & crowd & 276 \\
cSQM ZhEn & 1 & crowd & 70 \\
pSQM & 3 & professional & 6 \\
MQM & 3 & professional & 6 \\
\bottomrule
\end{tabular}
}
\caption{Details of all human evaluations.}
\label{tab:annotated_dataset}
\end{table}

To ensure maximum diversity in ratings for pSQM and MQM, we assigned documents in round-robin fashion to all 20 different sets of 3 raters from these pools. We chose an assignment order that roughly balanced the number of documents and segments per rater. Each rater was assigned a subset of documents, and annotated outputs from all 10 systems for those documents. Both documents and systems were anonymized and presented in a different random order to each rater. The number of segments per rater ranged from 6,830--7,220 for English$\to$German and from 9,860--10,210 for Chinese$\to$English.

\section{Results}

\subsection{Overall System Rankings}
\begin{figure}[!ht]
    \begin{minipage}{.48\textwidth}
    \includegraphics[width=1.03\textwidth]{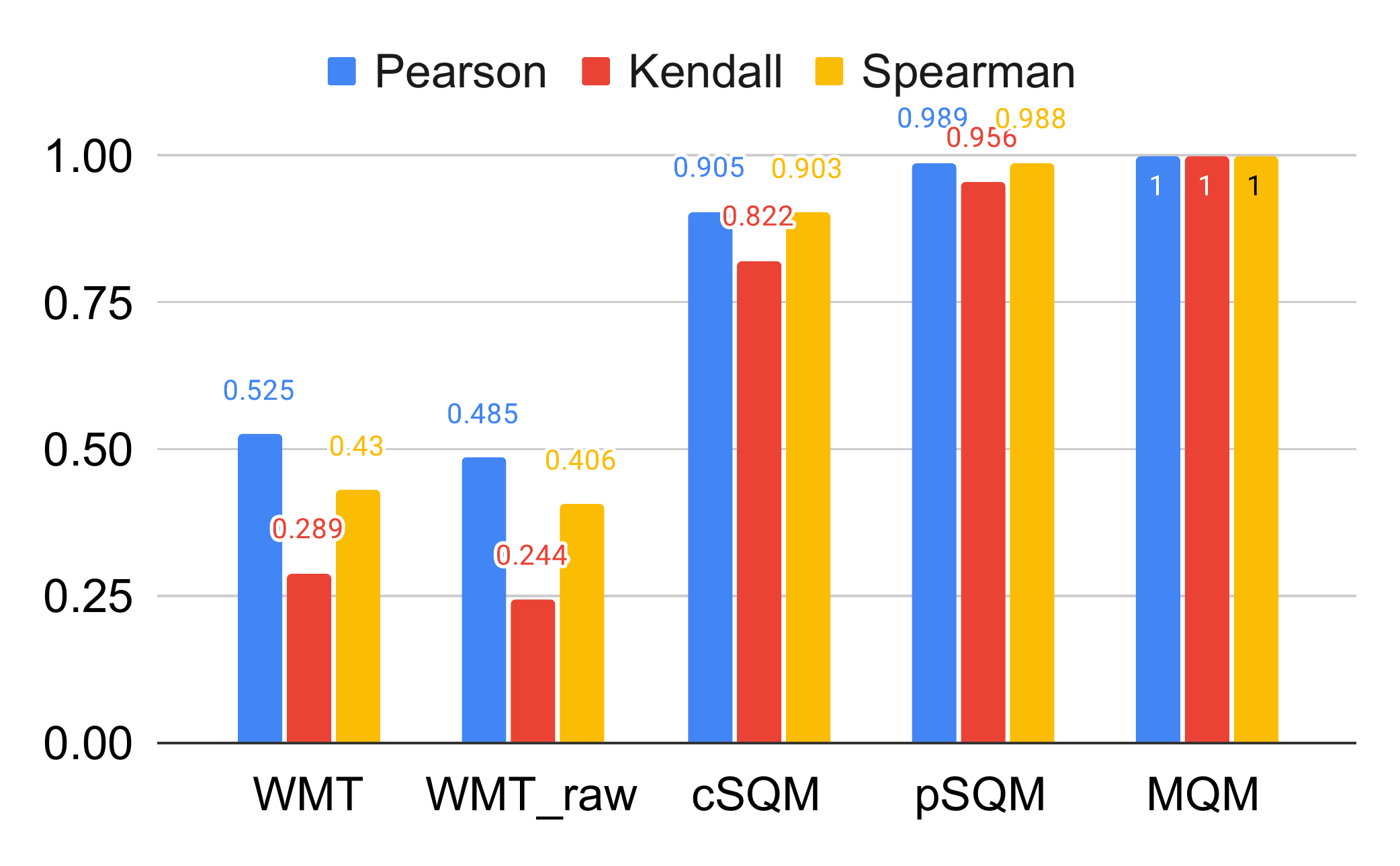}
    \vspace{-2.5em}
    \caption{English$\to$German: System correlation with the platinum ratings acquired with MQM.}
    \label{fig:ende-systemlevel-corr}
    \end{minipage}
\end{figure}

\begin{figure}[!ht]
    \begin{minipage}{.48\textwidth}
    \includegraphics[width=1.03\textwidth]{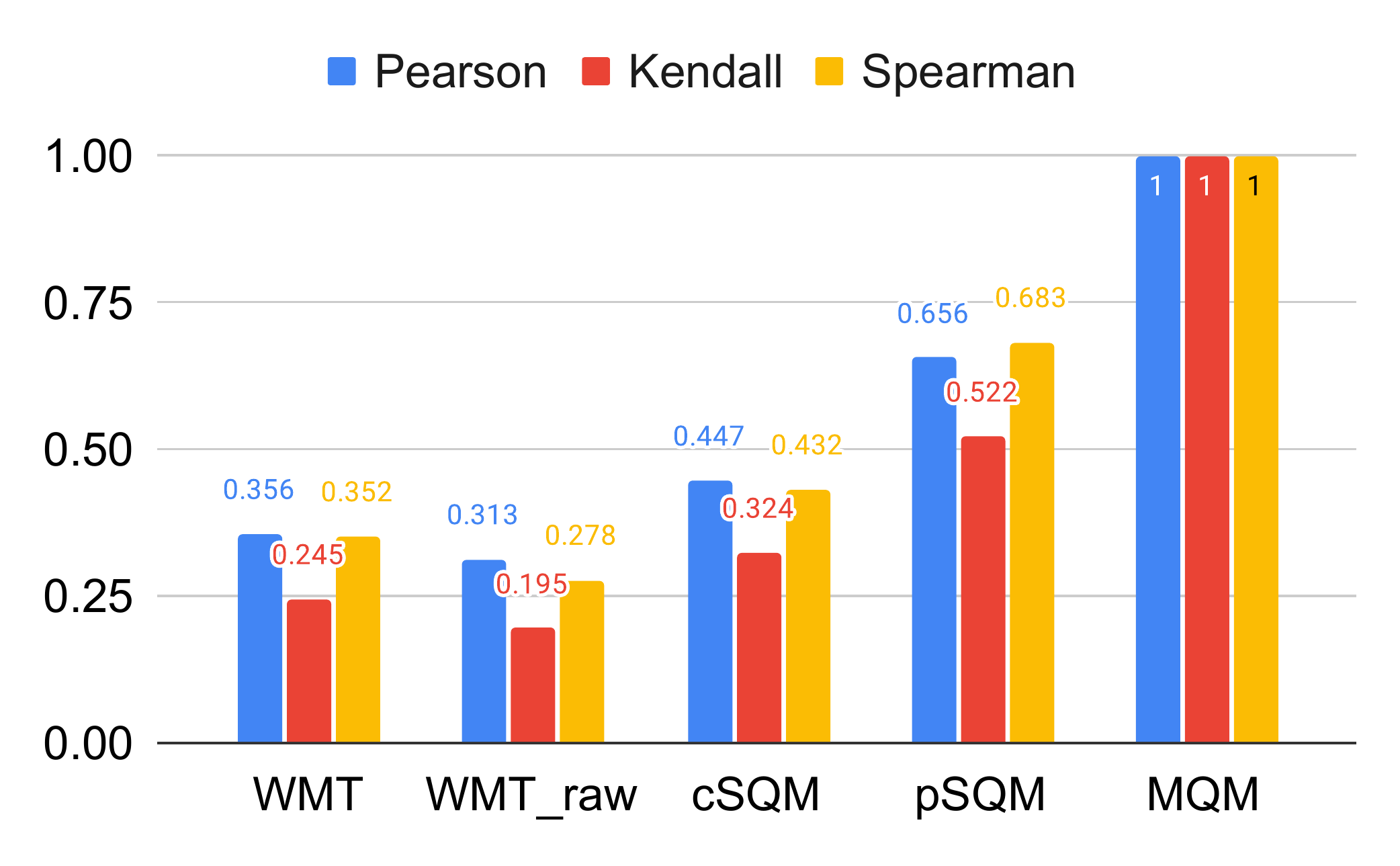}
    \vspace{-2.5em}
    \caption{English$\to$German: Segment correlation with the platinum ratings acquired with MQM.}
    \label{fig:ende-seglevel-corr}
    \end{minipage}
\end{figure}

\begin{figure}[!ht]
    \hfill\begin{minipage}{.48\textwidth}
    \includegraphics[width=1.03\textwidth]{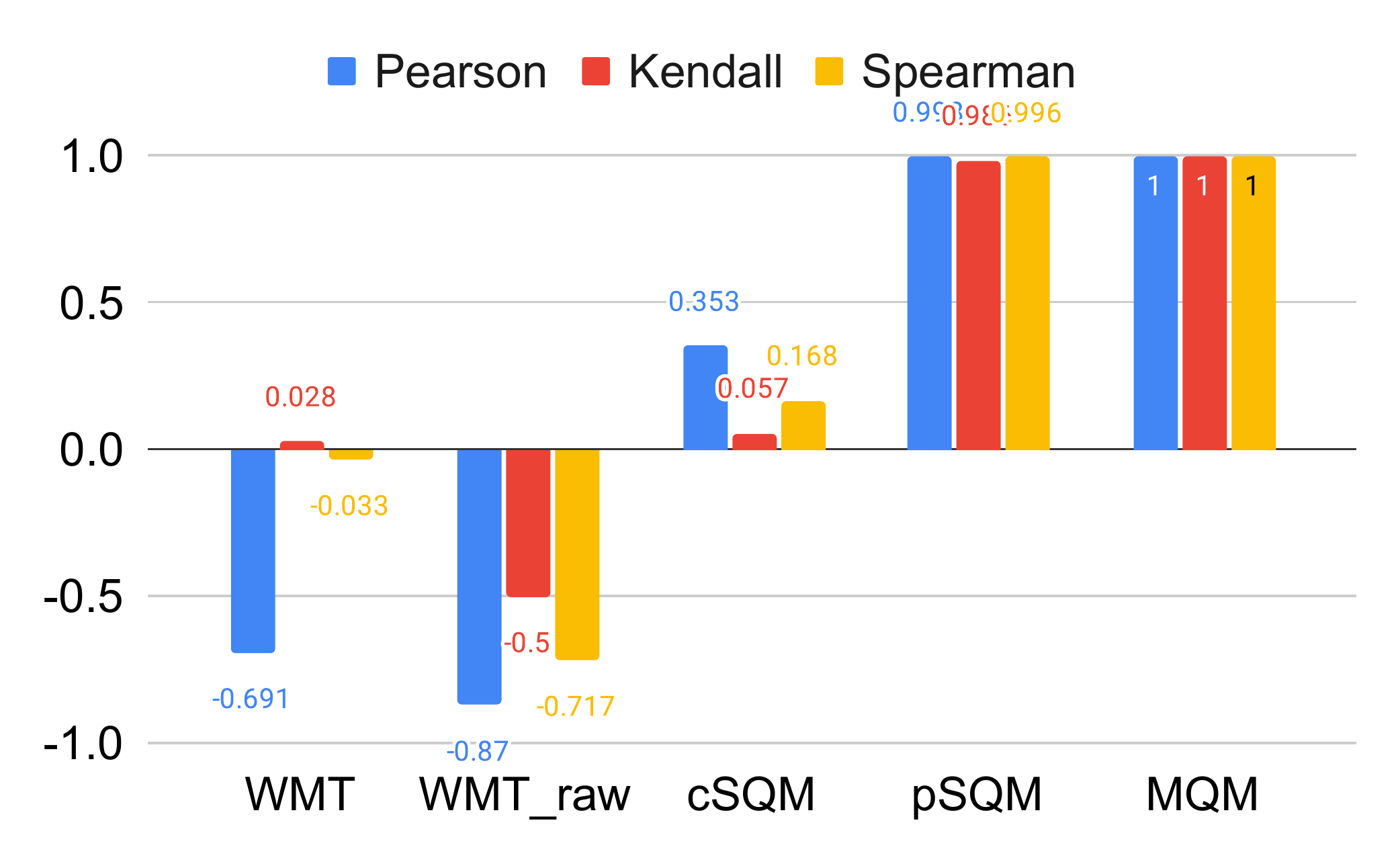}
    \vspace{-2.5em}
    \caption{Chinese$\to$English: System-level correlation with the platinum ratings acquired with MQM.}
    \label{fig:chen-systemlevel-corr}
     \end{minipage}
\end{figure}

\begin{figure}[!ht]
    \begin{minipage}{.48\textwidth}
    \includegraphics[width=1.03\textwidth]{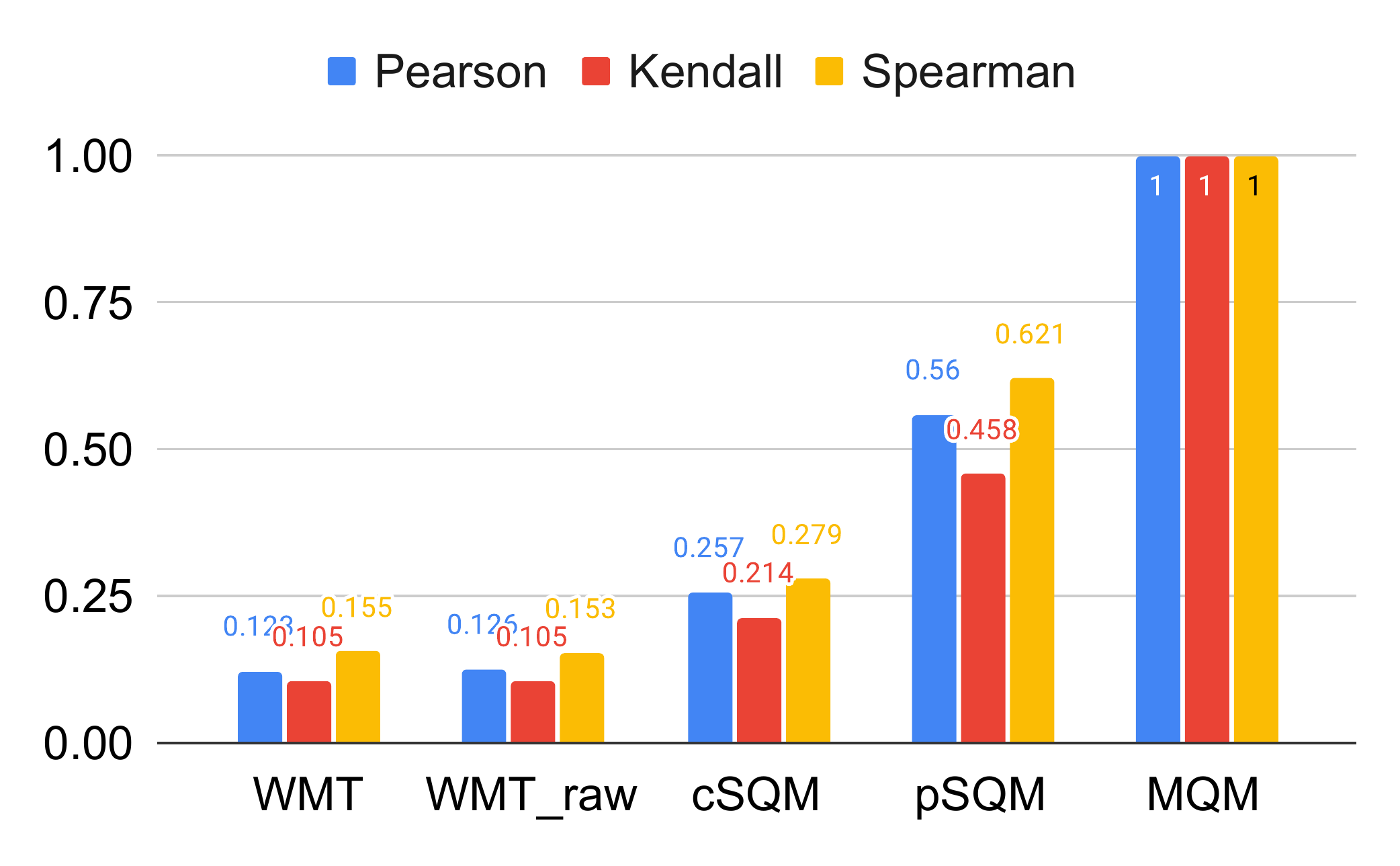}
    \vspace{-2.5em}
    \caption{Chinese$\to$English: Segment correlation with the platinum ratings acquired with MQM.}
    \label{fig:chen-seglevel-corr}
    \end{minipage}
\end{figure}

For each human evaluation setup, we calculate a system-level score by averaging the segment-level scores for each system. Results are summarized in Table~\ref{tab:ende_abs} (English$\to$German) and Table~\ref{tab:chen_abs} (Chinese$\to$English). The system- and segment-level correlations to our platinum MQM ratings are shown in Figure~\ref{fig:ende-systemlevel-corr} and \ref{fig:ende-seglevel-corr} (English$\to$German), and Figure~\ref{fig:chen-systemlevel-corr} and \ref{fig:chen-seglevel-corr} (Chinese$\to$English). Segment-level correlations are calculated only for segments that were evaluated by WMT. 
For both language pairs, we observe similar patterns when looking at the results of the different human evaluations and come to the following findings: 

\begin{table*}[!htb]\centering
\scalebox{0.77}{
\begin{tabular}{l|cccc|ccccc}\toprule
System & WMT$\uparrow$ & WMT RAW$\uparrow$ & cSQM$\uparrow$ & pSQM$\uparrow$ & MQM $\downarrow$ & Major$\downarrow$ & Minor$\downarrow$ & Fluency$\downarrow$ & Accuracy$\downarrow$\\ \midrule
Human-B & 0.569(1) & 90.5(1) & 5.31(1) & 5.16(1) & 0.75(1) & 0.22(1) & 0.54(1) & 0.28(1) & 0.47(1)\\
Human-A & 0.446(4) & 85.7(4) & 5.20(2) & 4.90(2) & 0.91(2) & 0.28(2) & 0.64(2) & 0.33(2) & 0.58(2)\\
Human-P & 0.299(10) & 84.2(9) & 5.04(5) & 4.32(3) & 1.41(3) & 0.57(3) & 0.85(3) & 0.50(3) & 0.91(3)\\
Tohoku-AIP-NTT & 0.468(3) & 88.6(2) & 5.11(3) & 3.95(4) & 2.02(4) & 0.94(4) & 1.14(4) & 0.61(5) & 1.40(4)\\
OPPO & 0.495(2) & 87.4(3) & 5.03(6) & 3.79(5) & 2.25(5) & 1.07(5) & 1.19(6) & 0.62(6) & 1.63(5)\\
eTranslation & 0.312(9) & 82.5(10) & 5.02(7) & 3.68(7) & 2.33(6) & 1.18(7) & 1.16(5) & 0.56(4) & 1.78(7)\\
Tencent\_Translation & 0.386(6) & 84.3(8) & 5.06(4) & 3.77(6) & 2.35(7) & 1.15(6) & 1.22(8) & 0.63(7) & 1.73(6)\\
Huoshan\_Translate & 0.326(7) & 84.6(6) & 5.00(8) & 3.65(8) & 2.45(8) & 1.23(8) & 1.23(9) & 0.64(8) & 1.80(8)\\
Online-B & 0.416(5) & 84.5(7) & 4.95(9) & 3.60(9) & 2.48(9) & 1.34(9) & 1.20(7) & 0.64(9) & 1.84(9)\\
Online-A & 0.322(8) & 85.3(5) & 4.85(10) & 3.32(10) & 2.99(10) & 1.73(10) & 1.32(10) & 0.76(10) & 2.23(10)\\
\bottomrule
\end{tabular}
}
\caption{English$\to$German: Different human evaluations for 10 submissions of the WMT20 evaluation campaign.} \label{tab:ende_abs}
\end{table*}

\begin{table*}[!htb]\centering
\scalebox{0.77}{
\begin{tabular}{l|cccc|ccccc}\toprule
System & WMT$\uparrow$ & WMT RAW$\uparrow$ & cSQM$\uparrow$ & pSQM$\uparrow$ & MQM $\downarrow$ & Major$\downarrow$ & Minor$\downarrow$ & Fluency$\downarrow$ & Accuracy$\downarrow$\\ \midrule
Human-A & - & - & 5.09(2) & 4.34(1) & 3.43(1) & 2.71(1) & 0.74(1) & 0.91(1) & 2.52(1)\\
Human-B & -0.029(9) & 74.8(9) & 5.03(7) & 4.29(2) & 3.62(2) & 2.81(2) & 0.82(10) & 0.95(2) & 2.66(2)\\
VolcTrans & 0.102(1) & 77.47(5) & 5.04(5) & 4.03(3) & 5.03(3) & 4.26(3) & 0.79(6) & 1.31(7) & 3.71(3)\\
WeChat\_AI & 0.077(3) & 77.35(6) & 4.99(8) & 4.02(4) & 5.13(4) & 4.39(4) & 0.76(4) & 1.24(5) & 3.89(4)\\
Tencent\_Translation & 0.063(4) & 76.67(7) & 5.04(6) & 3.99(5) & 5.19(5) & 4.43(6) & 0.79(8) & 1.23(4) & 3.96(5)\\
OPPO & 0.051(7) & 77.51(4) & 5.07(4) & 3.99(5) & 5.20(6) & 4.41(5) & 0.81(9) & 1.23(3) & 3.97(6)\\
THUNLP & 0.028(8) & 76.48(8) & 5.11(1) & 3.98(7) & 5.34(7) & 4.61(7) & 0.75(3) & 1.27(6) & 4.07(9)\\ 
DeepMind & 0.051(6) & 77.96(1) & 5.07(3) & 3.97(8) & 5.41(8) & 4.67(8) & 0.75(2) & 1.38(8) & 4.02(7)\\ 
DiDi\_NLP & 0.089(2) & 77.63(3) & 4.91(9) & 3.95(9) & 5.48(9) & 4.73(9) & 0.77(5) & 1.43(9) & 4.05(8)\\ 
Online-B & 0`.06(5) & 77.77(2) & 4.83(10) & 3.89(10) & 5.85(10) & 5.08(10) & 0.79(7) & 1.51(10) & 4.34(10)\\
\bottomrule
\end{tabular}
}
\caption{Chinese$\to$English: Different human evaluations for 10 submissions of the WMT20 evaluation campaign.}
\vspace{-1em}
\label{tab:chen_abs}
\end{table*}

\noindent
\textbf{(i) Human translations are underestimated by crowd workers:}
Already in 2016, \newcite{hassan2018achieving} claimed human parity for news-translation for Chinese$\to$English. We confirm the findings of \newcite{Toral18,laubli2018has} that when human evaluation is conducted correctly, professional translators can discriminate between human and machine translations. All human translations are ranked first by both the pSQM and MQM evaluations for both language pairs. The gap between human translations and MT is even more visible when looking at the MQM ratings which sets the human translations first by a large margin, demonstrating that the quality difference between MT and human translation is still large.
Another interesting observation is the ranking of Human-P for English$\to$German. Human-P is a reference translation generated using the paraphrasing method of \cite{freitag2020bleu} which asked linguists to paraphrase existing reference translations as much as possible while also suggesting using synonyms and different sentence structures. Our results support the assumption that crowd-workers are biased to prefer literal, easy-to-rate translations and rank Human-P low. Professional translators on the other hand are able to see the correctness of the paraphrased translations and ranked them higher than any MT output. Similar to the standard human translations, the gap between Human-P and the MT systems is larger when looking at the MQM ratings. In MQM, raters have to justify their ratings by labelling the error spans which helps to avoid penalizing non-literal translations.

\noindent
\textbf{(ii) WMT has low correlation with MQM:} 
The human evaluation in WMT was conducted by crowd-workers (Chinese$\to$English) or a mix of researchers/translators (English$\to$German) during the WMT evaluation campaign. 
Further, different to all other evaluations in this paper, WMT conducted a reference-based/monolingual human evaluation for Chinese$\to$English in which the machine translation output was compared to a human-generated reference.
When comparing the system ranks based on WMT for both language pairs with the ones generated by MQM, we can see low correlation for English$\to$German (see Figure~\ref{fig:ende-systemlevel-corr}) and even negative correlation for Chinese$\to$English (see Figure~\ref{fig:chen-systemlevel-corr}). We also see very low segment-level correlation for both language pairs (see Figure~\ref{fig:ende-seglevel-corr} and Figure~\ref{fig:chen-seglevel-corr}). Later, we will also show that the correlation of SOTA automatic metrics are higher than the human ratings generated by WMT. The results at least question the reliability of the human ratings acquired by WMT.

\noindent
\textbf{(iii) pSQM has high system-level correlation with MQM:}
The results for both language pairs suggest that pSQM and MQM are of similar quality as their system rankings mostly agree. Nevertheless, when zooming into the segment-level correlations, we observe a much lower correlation of $\sim$0.5 based on Kendall tau for both language pairs. The difference of the two approaches is also visible in the absolute differences of the individual systems. For instance the submissions of DiDi\_NLP and Tencent\_Translation for Chinese$\to$English are close for pSQM (only 0.04 absolute difference). MQM on the other hand shows a larger difference of 0.19 points. When the quality of two systems gets closer, a more fine-grained evaluation schema like MQM is needed. This is also important when doing system development where the difference between two variations for two systems can be minor. Looking into the future when we get closer to human translation quality, MQM will be needed for reliable evaluation. On the other hand, pSQM seems to be sufficient for an evaluation campaign like WMT.

\noindent
\textbf{(iv) MQM results are mainly driven by major and accuracy errors:}
In Table~\ref{tab:ende_abs} and Table~\ref{tab:chen_abs}, we also show the MQM error scores only based on Major/Minor errors or only based on Fluency or Accuracy errors. Interestingly, the MQM score based on accuracy errors or based on Major errors gives us almost the same rank as the full MQM score. Later in the paper, we will see that the majority of major errors are accuracy errors. This suggests the quality of an MT system is still driven mostly by accuracy errors as most fluency errors are judged minor.

\subsection{Error Category Distribution}
MQM provides fine-grained error categories grouped under 4 main categories (accuracy, fluency, terminology and style). 
The absolute error counts for all 3 ratings for all 10 systems are shown in Tables~\ref{tab:ende:mqm_err_cat} and \ref{tab:zhen:mqm_err_cat}.
The error category Accuracy/Mistranslation is responsible for the majority of major errors for both language pairs. This suggests that the main problem of MT is still mistranslation of words or phrases. The absolute number of errors is much higher for Chinese$\to$English which demonstrates that this translation pair is more challenging than English$\to$German.

Table~\ref{tab:ende:mqm_err_cat} decomposes system and human MQM scores 
per category for English$\to$German. Human translations get lower error counts in all categories, except 
for additions. It seems that human translators might add tokens for fluency which are not supported by 
the source. Both systems and humans are mostly penalized by accuracy/mistranslation errors, 
but systems record 4x more error points in these categories. Similarly, sentences with more
than 5 major errors (non-translation) are much more frequent for systems ($\sim\-28$x the human rate).
The best systems are quite different across categories. Tohoku is average in fluency but
outstanding in accuracy, eTranslation is excellent in fluency but worse in accuracy, and OPPO ranks
between the two other systems for both aspects. Compared to humans, the best systems are mostly penalized for
mistranslations and non-translation (badly garbled sentences).

Table~\ref{tab:zhen:mqm_err_cat} shows that the Chinese$\to$English translation task is more difficult than English$\to$German translation, with higher MQM error scores for human translations. Again, humans are performing better than systems across 
all categories except for additions, omissions and spelling. Many spelling mistakes relate to name formatting and capitalization which
is difficult for this language pair (see name formatting errors). Additions and omissions again highlight that humans might be ready
to compromise accuracy for fluency in some cases. Mistranslation and name formatting are the categories where the systems are penalized
the most compared to humans. When comparing systems, the differences between the best systems is less pronounced
than for English$\to$German, both in term of aggregate score and per-category counts.

\begin{table*}[!htb]\centering
\scalebox{0.80}{
\begin{tabular}{l |r|r||r | r r|| r r | r r |r r}\toprule
Error Categories & Errors & Major & Human  & \multicolumn{2}{c||}{All MT}  & \multicolumn{2}{c|}{Tohoku}    & \multicolumn{2}{c|}{OPPO}  & \multicolumn{2}{c}{eTrans}\\
                 & (\%) & (\%) & MQM  & MQM & vs H.  & MQM & vs H.  & MQM & vs H.  & MQM & vs H. 
\\\midrule
Accuracy/Mistranslation   & 33.2  & 27.2  & 0.296  & 1.285 & {\it 4.3} & 1.026 & {\it 3.5} & 1.219 & {\it 4.1} & 1.244 & {\it 4.2}\\
Style/Awkward             & 14.6  & 4.6  & 0.146  & 0.299 & {\it 2.0} & 0.289 & {\it 2.0} & 0.315 & {\it 2.1} & 0.296 & {\it 2.0}\\
Fluency/Grammar           & 10.7  & 4.7  & 0.097  & 0.224 & {\it 2.3} & 0.193 & {\it 2.0} & 0.215 & {\it 2.2} & 0.196 & {\it 2.0}\\
Accuracy/Omission         & 3.6  & 13.4  & 0.070  & 0.091 & {\it 1.3} & 0.063 & {\it 0.9} & 0.063 & {\it 0.9} & 0.120 & {\it 1.7}\\
Accuracy/Addition         & 1.8  & 6.7  & 0.067  & 0.025 & {\it 0.4} & 0.018 & {\it 0.3} & 0.024 & {\it 0.4} & 0.021 & {\it 0.3}\\
Terminology/Inappropriate & 8.3  & 7.0  & 0.061  & 0.193 & {\it 3.2} & 0.171 & {\it 2.8} & 0.189 & {\it 3.1} & 0.193 & {\it 3.2}\\
Fluency/Spelling          & 2.3  & 1.2  & 0.030  & 0.039 & {\it 1.3} & 0.030 & {\it 1.0} & 0.039 & {\it 1.3} & 0.028 & {\it 0.9}\\
Accuracy/Untranslated tex & 3.1  & 14.9  & 0.024  & 0.090 & {\it 3.8} & 0.082 & {\it 3.5} & 0.066 & {\it 2.8} & 0.098 & {\it 4.2}\\
Fluency/Punctuation       & 20.3  & 0.2  & 0.014  & 0.039 & {\it 2.8} & 0.067 & {\it 4.9} & 0.013 & {\it 1.0} & 0.011 & {\it 0.8}\\
Other                     & 0.5  & 5.2  & 0.005  & 0.010 & {\it 1.9} & 0.009 & {\it 1.6} & 0.010 & {\it 1.9} & 0.007 & {\it 1.2}\\
Fluency/Register          & 0.6  & 5.0  & 0.005  & 0.014 & {\it 3.0} & 0.009 & {\it 1.9} & 0.015 & {\it 3.2} & 0.015 & {\it 3.3}\\
Terminology/Inconsistent  & 0.3  & 0.0  & 0.004  & 0.005 & {\it 1.2} & 0.004 & {\it 0.9} & 0.005 & {\it 1.2} & 0.005 & {\it 1.2}\\
Non-translation           & 0.2  & 100.0  & 0.003  & 0.083 & {\it 28.3} & 0.041 & {\it 14.0} & 0.065 & {\it 22.0} & 0.094 & {\it 32.0}\\
Fluency/Inconsistency     & 0.1  & 1.3  & 0.003  & 0.002 & {\it 0.7} & 0.001 & {\it 0.3} & 0.001 & {\it 0.3} & 0.003 & {\it 1.0}\\
Fluency/Character enc.    & 0.1  & 3.7  & 0.002  & 0.001 & {\it 0.7} & 0.002 & {\it 1.0} & 0.001 & {\it 0.6} & 0.000 & {\it 0.2}\\
\midrule All accuracy     & 41.7  & 24.2  & 0.457  & 1.492 & {\it 3.3} & 1.189 & {\it 2.6} & 1.372 & {\it 3.0} & 1.483 & {\it 3.2}\\
All fluency               & 34.2  & 1.8  & 0.150  & 0.320 & {\it 2.1} & 0.303 & {\it 2.0} & 0.284 & {\it 1.9} & 0.253 & {\it 1.7}\\
All except acc. \& fluenc & 24.2  & 6.0  & 0.222  & 0.596 & {\it 2.7} & 0.526 & {\it 2.4} & 0.591 & {\it 2.7} & 0.596 & {\it 2.7}\\
\midrule All categories   & 100.0  & 12.1  & 0.829  & 2.408 & {\it 2.9} & 2.017 & {\it 2.4} & 2.247 & {\it 2.7} & 2.332 & {\it 2.8}\\
\bottomrule
\end{tabular}
}
\caption{
Category breakdown of MQM scores for English$\to$German
for human translations (A, B), machine translations (all systems) 
and some of the best systems (Tohohku, OPPO, eTranslation).
The ratio of system over human scores is in italics. 
Errors (\%) report the fraction of the total error counts in a category, 
Major (\%) report the fraction of major error for each category.} 
\label{tab:ende:mqm_err_cat}
\end{table*}

\begin{table*}[!htb]\centering
\scalebox{0.80}{
\begin{tabular}{l |r|r||r | r r|| r r | r r |r r}\toprule
Error Categories & Errors & Major & Human  & \multicolumn{2}{c||}{All MT}  & \multicolumn{2}{c|}{VolcTrans}    & \multicolumn{2}{c|}{WeChat}  & \multicolumn{2}{c}{Tencent}\\
                 & (\%) & (\%) & MQM  & MQM & vs H.  & MQM & vs H.  & MQM & vs H.  & MQM & vs H. 
\\\midrule
Accuracy/Mistranslation             & 42.2  & 71.5  & 1.687  & 3.218 & {\it 1.9} & 2.974 & {\it 1.8} & 3.108 & {\it 1.8} & 3.157 & {\it 1.9}\\
Accuracy/Omission                   & 8.6  & 61.3  & 0.646  & 0.505 & {\it 0.8} & 0.468 & {\it 0.7} & 0.534 & {\it 0.8} & 0.547 & {\it 0.8}\\
Fluency/Grammar                     & 13.8  & 18.4  & 0.381  & 0.442 & {\it 1.2} & 0.414 & {\it 1.1} & 0.392 & {\it 1.0} & 0.425 & {\it 1.1}\\
Locale/Name format                  & 6.4  & 74.5  & 0.250  & 0.505 & {\it 2.0} & 0.506 & {\it 2.0} & 0.491 & {\it 2.0} & 0.433 & {\it 1.7}\\
Terminology/Inappropriate           & 5.1  & 31.1  & 0.139  & 0.221 & {\it 1.6} & 0.220 & {\it 1.6} & 0.217 & {\it 1.6} & 0.202 & {\it 1.5}\\
Style/Awkward                       & 5.7  & 17.1  & 0.122  & 0.182 & {\it 1.5} & 0.193 & {\it 1.6} & 0.180 & {\it 1.5} & 0.185 & {\it 1.5}\\
Accuracy/Addition                   & 0.9  & 40.2  & 0.110  & 0.025 & {\it 0.2} & 0.017 & {\it 0.1} & 0.013 & {\it 0.1} & 0.018 & {\it 0.2}\\
Fluency/Spelling                    & 3.6  & 5.1  & 0.107  & 0.071 & {\it 0.7} & 0.071 & {\it 0.7} & 0.059 & {\it 0.6} & 0.073 & {\it 0.7}\\
Fluency/Punctuation                 & 11.1  & 1.4  & 0.028  & 0.035 & {\it 1.2} & 0.035 & {\it 1.3} & 0.031 & {\it 1.1} & 0.033 & {\it 1.2}\\
Locale/Currency format              & 0.4  & 8.8  & 0.011  & 0.010 & {\it 0.9} & 0.010 & {\it 0.9} & 0.010 & {\it 0.9} & 0.010 & {\it 0.9}\\
Fluency/Inconsistency               & 0.8  & 27.5  & 0.011  & 0.036 & {\it 3.3} & 0.028 & {\it 2.7} & 0.026 & {\it 2.4} & 0.038 & {\it 3.5}\\
Fluency/Register                    & 0.4  & 6.5  & 0.008  & 0.008 & {\it 1.0} & 0.008 & {\it 0.9} & 0.008 & {\it 1.0} & 0.009 & {\it 1.1}\\
Locale/Address format               & 0.3  & 65.7  & 0.008  & 0.025 & {\it 3.3} & 0.036 & {\it 4.7} & 0.033 & {\it 4.3} & 0.015 & {\it 2.0}\\
Non-translation                     & 0.0  & 100.0  & 0.006  & 0.024 & {\it 3.9} & 0.021 & {\it 3.3} & 0.012 & {\it 2.0} & 0.029 & {\it 4.7}\\
Terminology/Inconsistent            & 0.3  & 16.1  & 0.004  & 0.008 & {\it 2.3} & 0.007 & {\it 1.8} & 0.004 & {\it 1.2} & 0.010 & {\it 2.8}\\
Other                               & 0.1  & 4.1  & 0.003  & 0.003 & {\it 0.9} & 0.005 & {\it 1.7} & 0.002 & {\it 0.6} & 0.001 & {\it 0.4}\\
\midrule All accuracy               & 51.7  & 69.3  & 2.444  & 3.748 & {\it 1.5} & 3.463 & {\it 1.4} & 3.655 & {\it 1.5} & 3.721 & {\it 1.5}\\
All fluency                         & 29.8  & 10.5  & 0.535  & 0.593 & {\it 1.1} & 0.557 & {\it 1.0} & 0.517 & {\it 1.0} & 0.580 & {\it 1.1}\\
All except acc. \& fluency          & 18.5  & 41.7  & 0.546  & 0.986 & {\it 1.8} & 1.005 & {\it 1.8} & 0.955 & {\it 1.7} & 0.891 & {\it 1.6}\\
\midrule All categories             & 100.0  & 46.7  & 3.525  & 5.327 & {\it 1.5} & 5.025 & {\it 1.4} & 5.127 & {\it 1.5} & 5.192 & {\it 1.5}\\
\bottomrule
\end{tabular}
}
\caption{
Category breakdown of MQM scores for Chinese$\to$English
for human translations (A, B), machine translations (all systems) 
and some of the best systems (VolcTrans, WeChat, Tencent).
The ratio of system over human scores is in italics.
Errors (\%) report the fraction of the total error counts in a category, 
Major (\%) report the fraction of major error for each category.}  \label{tab:zhen:mqm_err_cat}
\end{table*}

\begin{table*}[!htb]\centering
\scalebox{0.80}{
\begin{tabular}{l |r r| r r | r r |r r| r r | r r}
\multicolumn{13}{c}{(a) English$\to$German}\\
\toprule
{Categories} & 
\multicolumn{2}{c|}{Rater 1} & 
\multicolumn{2}{c|}{Rater 2} & 
\multicolumn{2}{c|}{Rater 3} & 
\multicolumn{2}{c|}{Rater 4} & 
\multicolumn{2}{c|}{Rater 5} & 
\multicolumn{2}{c}{Rater 6}
\\
         & MQM & vs avg.  & MQM & vs avg.  & MQM & vs avg.  & MQM & vs avg.  & MQM & vs avg.  & MQM & vs avg.  \\
\midrule
Accuracy & 1.02 & {\it 0.84} & 0.82 & {\it 0.68} & 1.55 & {\it 1.28} & 1.42 & {\it 1.18} & 1.23 & {\it 1.02} & 1.21 & {\it 1.00}\\
Fluency  & 0.26 & {\it 0.96} & 0.34 & {\it 1.27} & 0.32 & {\it 1.18} & 0.28 & {\it 1.04} & 0.19 & {\it 0.70} & 0.23 & {\it 0.86}\\
Others   & 0.41 & {\it 0.80} & 0.63 & {\it 1.23} & 0.59 & {\it 1.14} & 0.57 & {\it 1.10} & 0.57 & {\it 1.10} & 0.32 & {\it 0.63}\\
\midrule
All      & 1.69 & {\it 0.85} & 1.79 & {\it 0.90} & 2.45 & {\it 1.23} & 2.27 & {\it 1.14} & 1.98 & {\it 1.00} & 1.76 & {\it 0.88}\\
\bottomrule
\end{tabular}
}

\scalebox{0.80}{
\begin{tabular}{l |r r| r r | r r |r r| r r | r r}
\multicolumn{13}{c}{(b) Chinese$\to$English}\\
\toprule
{Categories} & 
\multicolumn{2}{c|}{Rater 1} & 
\multicolumn{2}{c|}{Rater 2} & 
\multicolumn{2}{c|}{Rater 3} & 
\multicolumn{2}{c|}{Rater 4} & 
\multicolumn{2}{c|}{Rater 5} & 
\multicolumn{2}{c}{Rater 6}
\\
            & MQM & vs avg.  & MQM & vs avg.  & MQM & vs avg.  & MQM & vs avg.  & MQM & vs avg.  & MQM & vs avg.  \\
\midrule
Accuracy    & 3.34 & {\it 0.96} & 3.26 & {\it 0.94} & 3.31 & {\it 0.95} & 2.51 & {\it 0.72} & 4.57 & {\it 1.31} & 3.91 & {\it 1.12}\\
Fluency     & 0.39 & {\it 0.68} & 0.50 & {\it 0.87} & 1.13 & {\it 1.95} & 0.33 & {\it 0.57} & 0.59 & {\it 1.02} & 0.53 & {\it 0.92}\\
Others      & 0.70 & {\it 0.78} & 0.75 & {\it 0.83} & 0.85 & {\it 0.94} & 0.66 & {\it 0.74} & 1.11 & {\it 1.24} & 1.32 & {\it 1.47}\\\midrule
All         & 4.43 & {\it 0.89} & 4.51 & {\it 0.91} & 5.29 & {\it 1.07} & 3.50 & {\it 0.71} & 6.27 & {\it 1.26} & 5.76 & {\it 1.16}\\
\bottomrule
\end{tabular}
}
\caption{MQM per rater and category.
The ratio of a rater score over the average score is in italics.} \label{tab:mqm_err_cat_rater}
\end{table*}

\subsection{Document-error Distribution}
\begin{figure}[!ht]
    \begin{minipage}{.48\textwidth}
    \includegraphics[width=1.0\textwidth]{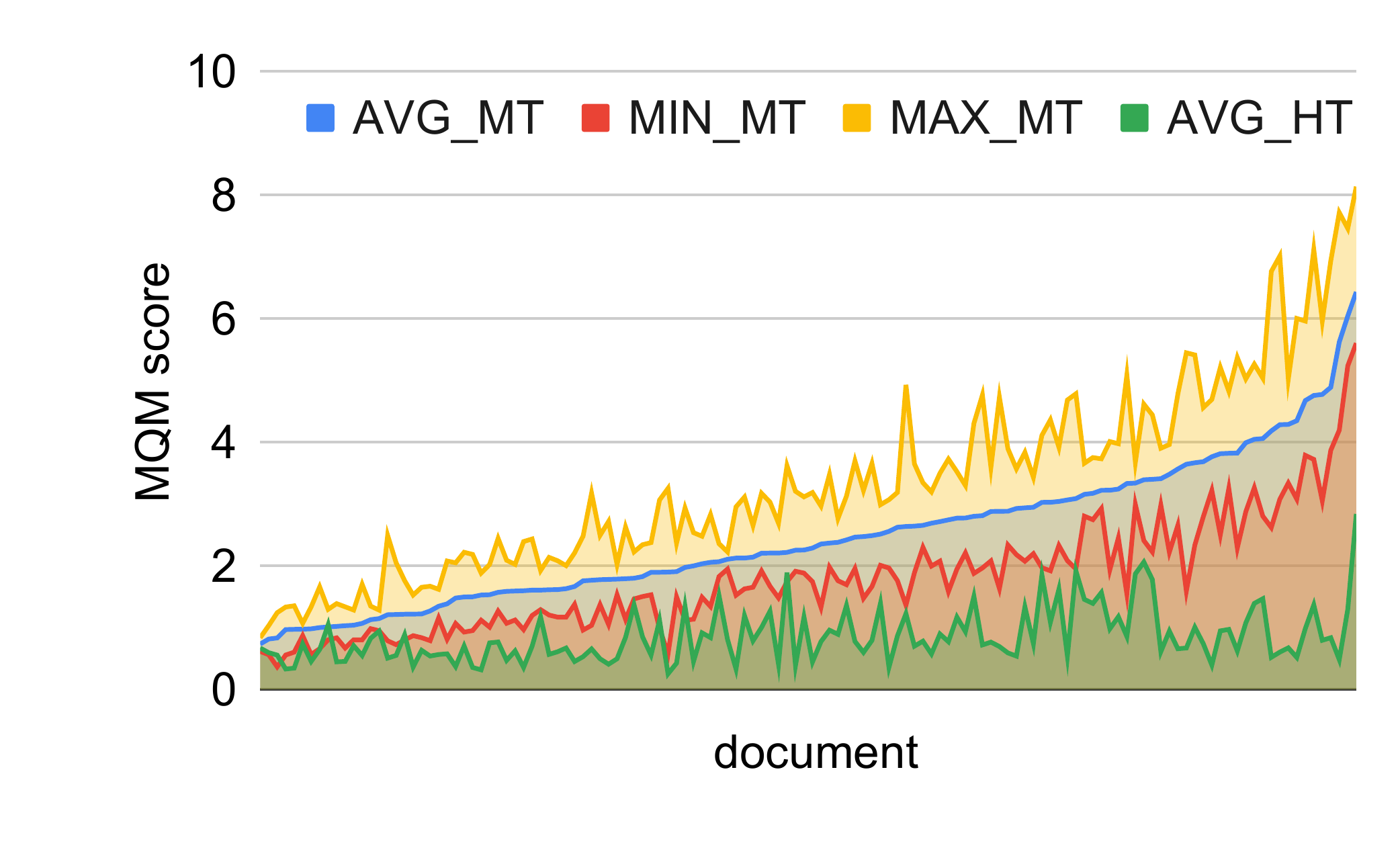}
    \vspace{-2em}
    \caption{EnDe: Document-level MQM scores.}

    \label{fig:ende-docdistr}
    \end{minipage}
    \vspace{-1em}
\end{figure}

We calculate document-level scores by averaging the segment level scores of each document. We show the average document scores of all MT systems and all human translations (HT) for English$\to$German in Figure~\ref{fig:ende-docdistr}. The translation quality of humans is very consistent over all documents and gets a MQM score of around 1 which is equivalent to one minor error. This demonstrates that the translation quality of humans is  consistent independent of the underlying source sentence.
The distribution of MQM errors for machine translations looks much different. For some documents, MT gets very close to human performance, while for other documents the gap is clearly visible. Interestingly, all MT systems have similar problems with the same subset of documents which demonstrated that the quality of MT output is more conditioned on the actual input sentence and not only on the underlying MT system.

\begin{figure}[!ht]
    \begin{minipage}{.48\textwidth}
    \includegraphics[width=1.0\textwidth]{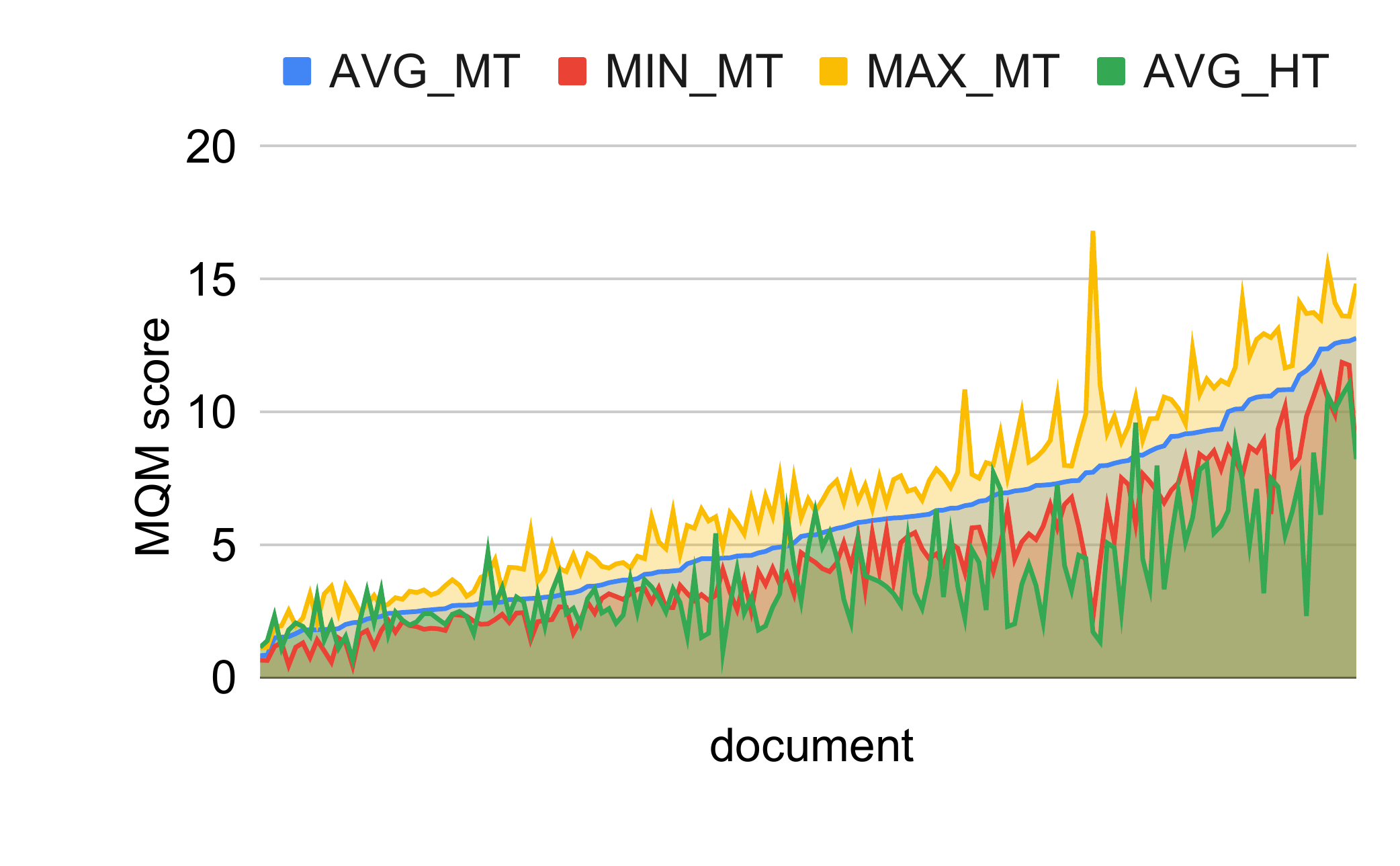}
    \vspace{-2em}
    \caption{ZhEn: Document-level MQM scores.}
    \label{fig:chen-docdistr}
    \end{minipage}
    \vspace{-1em}
\end{figure}

The MQM document-level scores for Chinese$\to$English are shown in Figure~\ref{fig:chen-docdistr}. The distribution of MQM errors for the MT output looks very similar to the ones for English$\to$German. There are documents that are more challenging for some MT systems than others. 
Although the document-level scores are mostly lower for human translations, the distribution looks similar to the ones from MT systems. 
We first suspected that the reference translations were post-edited from MT. 
This is not the case: these translations originate from professional translators without access to post-editing but with access to CAT tools (mem-source and translation memory).
Another possible explanation is the nature of the source sentences. Most sentences come from Chinese government news pages which have a formal style that may be difficult to render in English.

\subsection{Annotator Agreement and Reliability}

Our annotations were performed by professional raters with MQM training. All raters were given roughly the same amount of work, 
with the same number of segments from each system. This setup should result in similar aggregated rater scores.

 Table~\ref{tab:mqm_err_cat_rater}(a) reports the scores per rater aggregated over the main error categories for English$\to$German.
All raters provide scores within $\pm 20\%$ around the mean, with rater 3 being the most severe rater 
and rater 1 the most permissive. Looking at individual ratings, rater 2 rated fewer errors in accuracy categories but used the
style/awkward category more for errors outside of fluency/accuracy. Conversely, rater 6 barely used this category.
Differences in error rates among raters are not severe but could be reduced with corrections from an 
annotation model~\cite{paun-etal-2018-comparing}, especially when working with larger annotator pools.

The rater comparison on Chinese$\to$English in Table~\ref{tab:mqm_err_cat_rater}(b) reports a wider range of scores than for English$\to$German.
All raters provide scores within $\pm 30\%$ around the mean. This difference might be due to the greater difficulty of
the translation task itself introducing more ambiguity in the labeling. In the future, it would be interesting to compare
if translation between languages of different families suffer larger annotator disagreement for MQM ratings.

\subsection{Number of MQM Ratings Required}
\begin{figure}[!ht]
    \begin{minipage}{.48\textwidth}
    \includegraphics[width=1.0\textwidth]{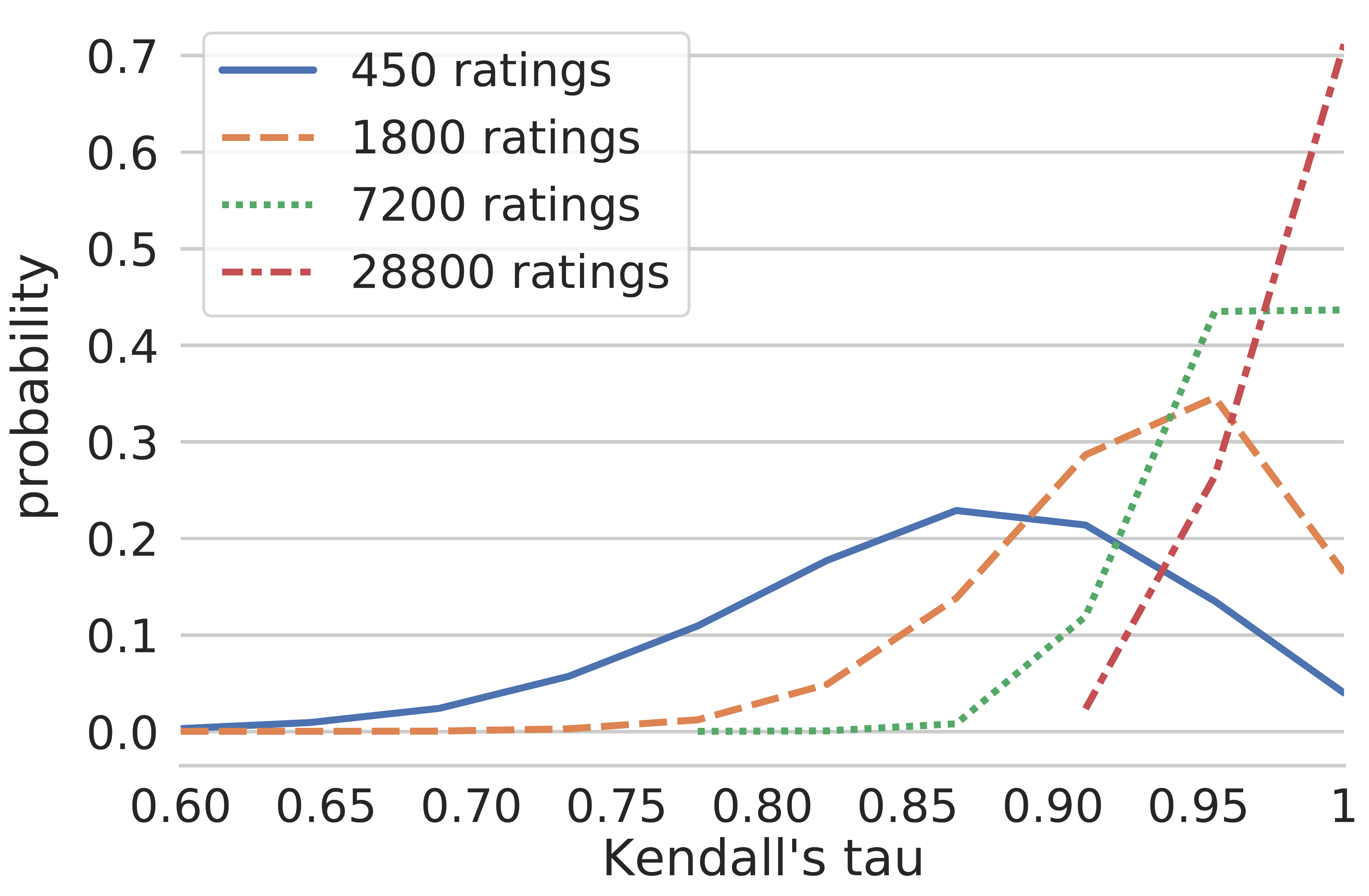}
    \vspace{-2em}
    \caption{Distributions of Kendall's $\tau$ of system level scores for English$\to$German. As the number of ratings increases, the distribution of Kendall's $\tau$ converges to the Dirac distribution at 1. All systems use 1 rater per sentence and 3 consecutive sentences per document. The width of $95\%$ CI is small (< 0.02), and thus is not shown here.}
    \label{fig:kendall-tau-various-num-ratings-mqm}
    \end{minipage}
\end{figure}

Human evaluation with professional translators is more expensive than using the crowd. To keep the cost as low as possible, we compute the minimum number of ratings required to get a reliable human evaluation.
We simulate new MQM rating projects by bootstrapping from the existing MQM data.\footnote{To make the bootstrapping more efficient, we computed the covariance matrix of the MQM ratings of all translation systems, and bootstrapped from a multi-variate Gaussian.} We compute Kendall's $\tau$ correlation of the simulated system level scores with the system level scores obtained from the full MQM data set. Note that later should be considered as the ground truth when estimating the accuracy of simulated MQM projects. 
See Figure~\ref{fig:kendall-tau-various-num-ratings-mqm} for the change of distributions of Kendall's $\tau$ for English$\to$German as the number of ratings increases.

\begin{figure}[!ht]
    \begin{minipage}{.48\textwidth}
    \includegraphics[width=1.0\textwidth]{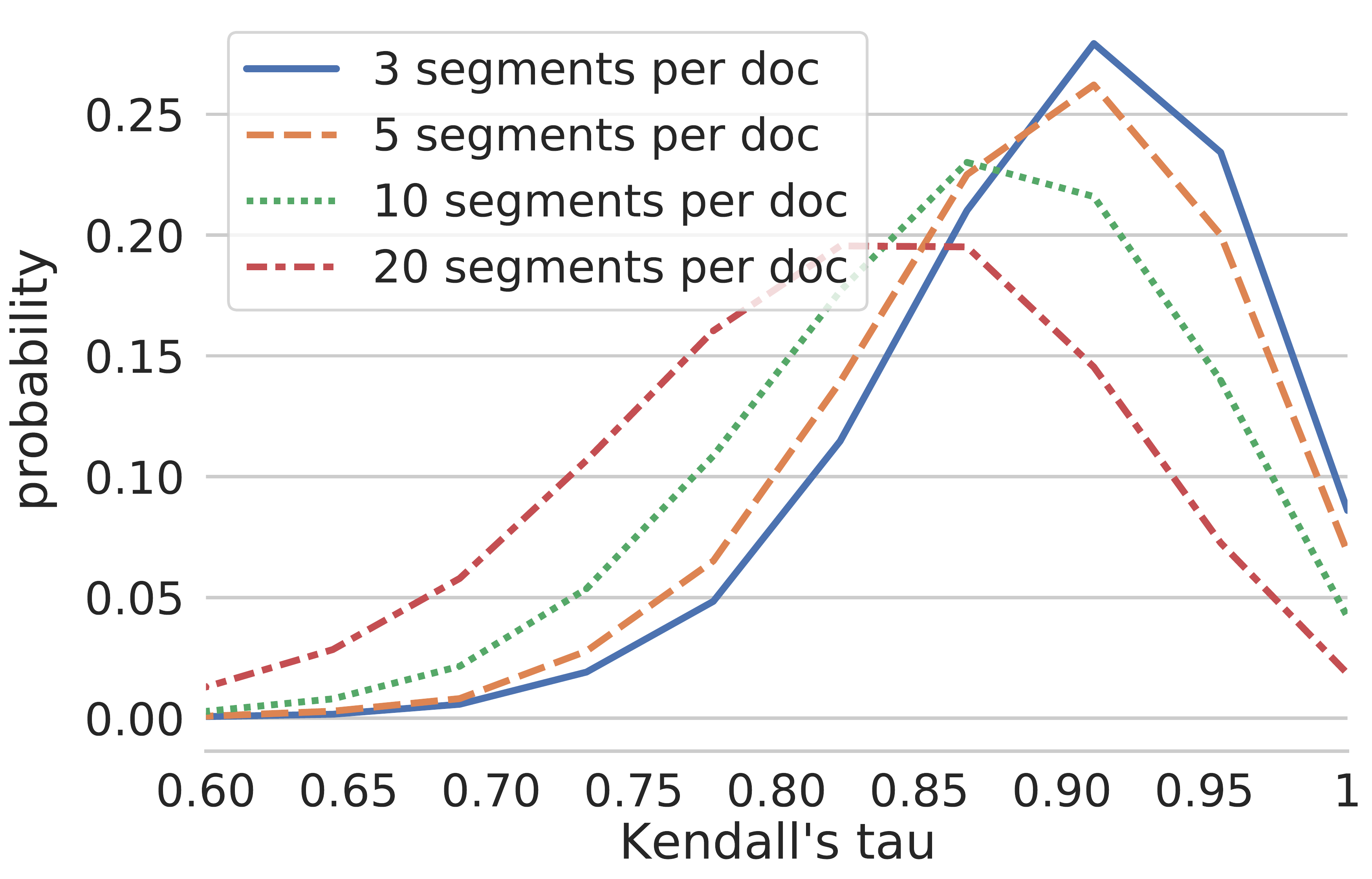}
    \vspace{-2em}
    \caption{System-level Kendall's $\tau$ for different distribution schema of 900 segment-level ratings for English$\to$German.}
    \label{fig:kendall-tau-900-mqm}
    \end{minipage}
\end{figure}

Figure~\ref{fig:kendall-tau-900-mqm} shows the effect of different distributing schema for a fixed budget of 900 segment-level ratings. The system level scores become more accurate when limiting the number of segment-level ratings to 3 consecutive sentences in each document and thus distributing the 900 segment-level scores over more documents.

Once the items to be rated is fixed for one system, aligning the ratings across different systems makes the comparison of two system more accurate. For MQM, this means that to compare different systems, it helps to rate the same documents, and the same sentences in the corresponding documents. When possible, using the same rater(s) to rate the corresponding sentences for different systems further improves the accuracy of the comparison between systems.

Finally, we estimate the number of ratings needed for MQM on different language pairs. The estimations are for systems with 3 consecutive sentences rated per document, and 1 rating per sentence. We further align the documents and the sentences rated across systems, but we do not align raters for corresponding sentences. We estimate the minimum number of ratings required such that the expected Kendall's $\tau$ correlation with the full data set $\ge 0.9$.

\begin{table}[!htb]\centering
\scalebox{0.80}{
\begin{tabular}{c|c}\toprule
language pair & number of ratings required \\
\midrule
English$\to$German & 951 \\
Chinese$\to$English  & 3720 \\
\bottomrule
\end{tabular}
}
\caption{MQM: Number of required ratings per system to achieve Kendall's $\tau$ of 0.9}
\label{tab:ratings-required}
\end{table}

\subsection{Impact on Automatic Evaluation}

\begin{figure*}[htb]
    \centering
    \includegraphics[scale=0.36]{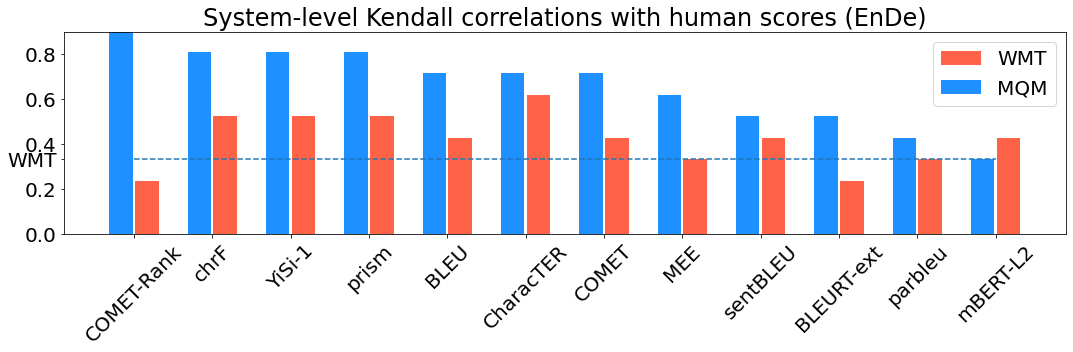}
    \includegraphics[scale=0.36]{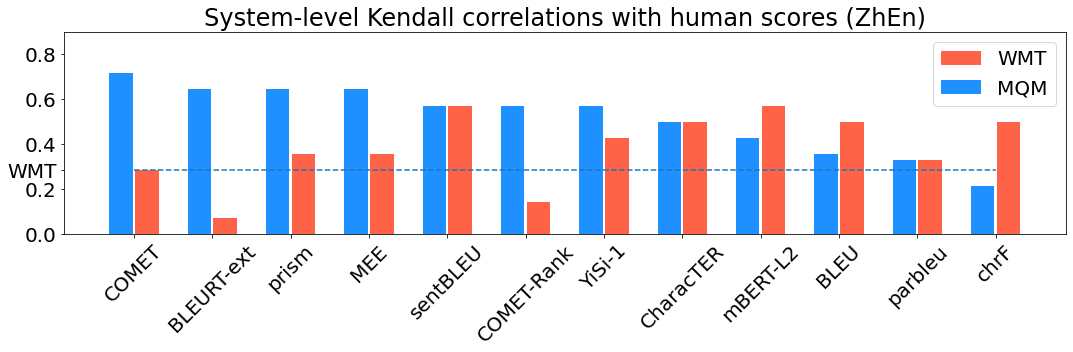}
    \caption{System-level metric performance with MQM and WMT scoring for: (a) EnDe, top panel; and (b) ZhEn, bottom panel. The horizontal blue line indicates the correlation between MQM and WMT human scores.}
    \label{fig:metrics}
\end{figure*}

We compared the performance of automatic metrics submitted to the WMT20 Metrics Task
when gold scores came from the original WMT ratings to the performance when gold scores were derived from our MQM ratings.
Figure~\ref{fig:metrics} shows Kendall's tau correlation for selected metrics at the system level for English$\to$German and Chinese$\to$English;\footnote{The official WMT system-level results use Pearson correlation, but since we are rating fewer systems (only 7 in the case of EnDe), Kendall is more meaningful; it also corresponds more directly to the main use case of system ranking.} full results are in Appendix~\ref{sec:full-metrics}. As would be expected from the low correlation between MQM and WMT scores, the ranking of metrics changes completely under MQM. In general, metrics that are not solely based on surface characteristics do somewhat better, though this pattern is not consistent (for example, chrF has a correlation of 0.8 for EnDe). Metrics tend to correlate better with MQM than they do with WMT, and almost all achieve better MQM correlation than WMT does (horizontal dotted line).

\begin{table}[!htb]\centering
\scalebox{0.80}{
\begin{tabular}{l|rr|rr}\toprule
\multicolumn{1}{c|}{Average} & \multicolumn{2}{c|}{EnDe} & \multicolumn{2}{c}{ZhEn} \\
\multicolumn{1}{c|}{correlations} & WMT & MQM & WMT & MQM \\
\midrule
Pearson, sys-level & 0.539 & 0.883 & 0.318 & 0.551 \\
        & \it 0.23 & \it 0.02 & \it 0.41 & \it 0.21 \\
Kendall, sys-level & 0.436 & 0.637 & 0.309 & 0.443 \\
        & \it 0.27 & \it 0.10 & \it 0.42 & \it 0.23 \\ 
Kendall, sys-level, & 0.467 & 0.676 & 0.514 & 0.343 \\
~~baseline metrics         & \it 0.20 & \it 0.06 & \it 0.10 & \it 0.34 \\                    
Kendall, sys-level, & 0.387 & 0.123 & 0.426 & 0.159 \\
 ~$+$ human     & \it 0.26 & \it 0.68 & \it 0.20 & \it 0.64 \\ 
\midrule
Kendall, seg-level & 0.170 & 0.228 & 0.159 & 0.298 \\
                    & \it 0.00 & \it 0.00 & \it 0.00 & \it 0.00 \\ 
Kendall, seg-level, & 0.159 & 0.161 & 0.157 & 0.276 \\
 ~$+$ human   & \it 0.00 & \it 0.00 & \it 0.00 & \it 0.00 \\  
\bottomrule
\end{tabular}
}
\caption{Average correlations for various subsets of metrics at different granularities. Numbers in italics are average p-values from two-tailed tests, indicating the probability that the observed correlation was due to chance.}
\label{tab:metrics}
\end{table}

Table~\ref{tab:metrics} shows average correlations with WMT and MQM gold scores for different subsets of metrics at different granularities. At the system level, correlations are higher for MQM than WMT, and for EnDe than ZhEn. Correlations to MQM are quite good, though on average they are statistically significant only for EnDe. Interestingly, the average performance of baseline metrics (BLEU, sentBLEU, TER, chrF, chrF++) is similar to the global average for all metrics in all conditions except for ZhEn WMT, where it is substantially better. Adding human translations\footnote{One additional standard reference and one paraphrased reference for EnDe, and one standard reference for ZhEn.} to the outputs scored by the metrics results in a large drop in performance, especially for MQM due to human outputs being rated unambiguously higher than MT by MQM. Segment-level correlations are generally much lower than system-level, though they are significant due to having greater support. MQM correlations are again higher than WMT at this granularity, and are higher for ZhEn than EnDe, reversing the pattern from system-level results and suggesting a potential for improved system-level metric performance through better aggregation of segment-level scores.

\section{Conclusion}
\label{sec:conclusion}
As part of this work, we proposed a standard MQM scoring scheme that is appropriate for high-quality MT. We used MQM to acquire ratings by professional translators for the recent WMT 2020 evaluation campaign for Chinese$\to$English and English$\to$German and used them as a platinum standard for comparison to different simpler evaluation methodologies and crowd worker evaluations. We release all ratings acquired in this study to encourage further research on this dataset for both human evaluation and automatic evaluation.

Our study shows that crowd-worker human evaluations (as conducted by WMT) have low correlation with MQM, and the resulting system-level rankings are quite different. This finding questions previous conclusions made on the basis of crowd-worker human evaluation, especially for high-quality MT. We further come to the surprising finding that many automatic metrics, and in particular embedding-based ones, already outperform crowd-worker human evaluation.
Unlike ratings acquired by crowd-worker and ratings acquired by professional translators on simpler human evaluation methodologies, MQM labels acquired with professional translators show a large gap between the quality of human and machine generated translations. This demonstrates that MT is still far from human parity. Furthermore, we characterize the current error types in human and machine translations, highlighting which error types are responsible for the difference between the two. We hope that researchers will use this as motivation to establish more error-type specific research directions.
Finally, we give recommendations of how many MQM labels are required to establish a reliable human evaluation and how these ratings should be distributed across documents.

\bibliography{tacl2018}
\bibliographystyle{acl_natbib}
\clearpage

\appendix

\newpage

\section{MQM Summary}
\label{sec:mqm-summary}

\begin{figure*}[tb]
    \centering
    \includegraphics[scale=0.45]{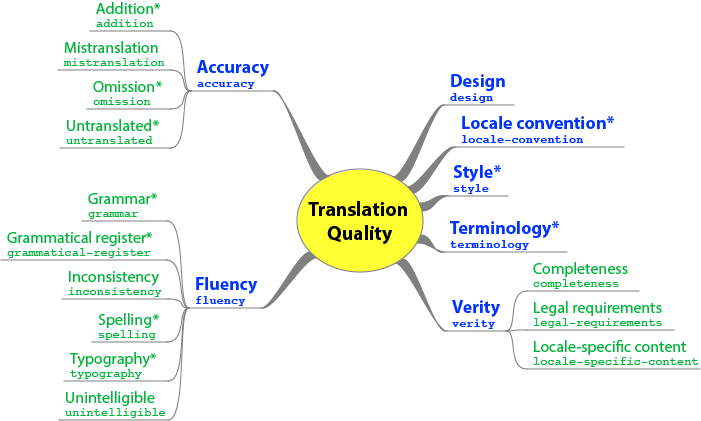}
    \caption{MQM Core issue hierarchy.}
    \label{fig:mqm-core}
\end{figure*}

The Multidimensional Quality Metrics (MQM) framework was developed in the EU QTLaunchPad and QT21 projects (2012--2018) \href{www.qt21.eu}{(www.qt21.eu)}.
It provides a generic methodology for assessing translation quality that can be adapted to a wide range of evaluation needs. The central idea is to establish a standard hierarchy of translation {\em issues} (potential errors) that can be pruned or extended with new issues as required. Annotators identify issues in text at a suitable granularity, and the results are summarized using a procedure that is specific to the application.

The MQM standard
\href{http://www.qt21.eu/mqm-definition}{(www.qt21.eu/mqm-definition)}
consists of a controlled vocabulary for describing issues, a scoring mechanism for aggregating annotation results, an XML formalism for describing specific {\em  metrics} (instantiations of MQM), a set of guidelines for selecting issues, and mappings from legacy metrics to MQM. All components except the vocabulary and XML mechanism are considered suggestive, and may be modified as required. Figure~\ref{fig:mqm-core} depicts the MQM Core issue hierarchy, intended to cover common issues arising in translated texts.

Guidelines for adapting MQM to scientific research are provided in the standard, and augmented by
\href{http://qt21.eu/downloads/MQM-usage-guidelines.pdf}{(MQM-usage-guidelines.pdf)}. The main points can be summarized as follows:
\begin{itemize}
\item Choose an issue hierarchy suitable to the research questions being addressed, 
introducing new issues as needed,\footnote{These must not overlap semantically with issues in the controlled vocabulary.} and pruning irrelevant issues to reduce ambiguity and cognitive load. Specify the granularity of the text units to which the issues will apply; this may range from sub-sentential spans to multi-document collections.

\item If possible, use expert human translators or translators to perform annotations; three annotators per text item is recommended. Provide training in the use of the annotation tool, and guidelines for interpreting the issue hierarchy. These may be augmented with examples or decision trees, and a calibration set containing known errors can be used to assure annotator competence.

\item Annotation should proceed in short segments (30 minutes), and the allocated time should take text difficulty into account. Annotation cost is estimated to be approximately 1 USD / segment (assuming three annotators), but can be highly variable. Annotation within document context is assumed implicitly.

\item Analysis can produce aggregate scores or finer-grained summaries. 
The specification recommends that each issue be graded with a severity: none, minor, major, or critical.
Aggregate scores can weight each issue by type (the default is to weight all types equally) and
by  severity (recommended scores are 0, 1, 10, and 100, respectively).
\end{itemize}

\section{MQM for Broad-Coverage MT}
\label{sec:mqm-details}

\subsubsection*{Annotation}

\begin{table*}[htb]\centering
\scalebox{0.80}{
\begin{tabular}{ll|l}\toprule
\multicolumn{2}{l|}{Error Category} & Description \\
\midrule
Accuracy & Addition    & Translation includes information not present in the source. \\
    & Omission         & Translation is missing content from the source. \\
    & Mistranslation   & Translation does not accurately represent the source.\\
    & Untranslated text & Source text has been left untranslated. \\
\midrule
Fluency & Punctuation   & Incorrect punctuation (for locale or style). \\
    & Spelling          & Incorrect spelling or capitalization. \\
    & Grammar           & Problems with grammar, other than orthography. \\
    & Register          & Wrong grammatical register (eg, inappropriately informal pronouns). \\
    & Inconsistency     & Internal inconsistency (not related to terminology). \\
    & Character encoding          & Characters are garbled due to incorrect encoding. \\
\midrule
Terminology & Inappropriate for context & Terminology is non-standard or does not fit context.\\
            & Inconsistent use & Terminology is used inconsistently.\\
\midrule
Style & Awkward & Translation has stylistic problems.\\
\midrule
Locale & Address format & Wrong format for addresses.\\
convention  & Currency format & Wrong format for currency.\\
    & Date format & Wrong format for dates. \\
    & Name format & Wrong format for names. \\
    & Telephone format & Wrong format for telephone numbers. \\
    & Time format & Wrong format for time expressions. \\
\midrule
Other & & Any other issues. \\
\midrule
Source error & & An error in the source. \\
\midrule
Non-translation & & Impossible to reliably characterize the 5 most severe errors.\\
\bottomrule
\multicolumn{3}{c}{}\\
\end{tabular}
}
\caption{MQM hierarchy.}
\label{tab:mqm-hierarchy}
\end{table*}

Our broad-coverage MT issue hierarchy is shown in Table~\ref{tab:mqm-hierarchy}. It is intended to be applied at the segment level by annotators with access to document context. We based it loosely on the MQM core hierarchy, with modifications established in collaboration with expert translators from our rater pool who had MQM experience. After an initial pilot run, we added several sub-categories to {\em Locale convention} for the sake of consistency.\footnote{An alternative and arguably preferable strategy would have been to collapse all sub-categories for locale.} Apart from clarifying the definitions of some categories, our main change was to add a {\em Non-translation} category to cover situations where identifying individual errors would be meaningless. At most one Non-translation error can be assigned to a segment, and choosing Non-translation precludes the identification of other errors in that segment.

\begin{table*}[!htb]\centering
\scalebox{0.80}{
\begin{tabular}{p{0.09\textwidth}|p{\textwidth}}\toprule
Severity & Description \\
\midrule
Major & Errors that may confuse or mislead the reader due to significant change in meaning or because they appear in a visible or important part of the content. \\
\midrule
Minor & Errors that don't lead to loss of meaning and wouldn't confuse or mislead the reader but would be noticed, would decrease stylistic quality, fluency or clarity, or would make the content less appealing.\\
\midrule
Neutral & Use to log additional information, problems or changes to be made that don't count as errors, e.g. they reflect a reviewer’s choice or preferred style.\\
\bottomrule
\end{tabular}
}
\caption{MQM severity levels.}
\label{tab:mqm-severity}
\end{table*}

\begin{table*}[!htb]\centering
\scalebox{1.00}{
\noindent\fbox{%
\parbox{1.0\textwidth}{%
You will be assessing translations at the segment level, where a segment may contain one or more sentences. Each segment is aligned with a corresponding source segment, and both segments are displayed within their respective documents. Annotate segments in natural order, as if you were reading the document. You may return to revise previous segments.\\

Please identify all errors within each translated segment, up to a maximum of five. If there are more than five errors, identify only the five most severe. If it is not possible to reliably identify distinct errors because the translation is too badly garbled or is unrelated to the source, then mark a single {\em Non-translation} error that spans the entire segment.\\
 
To identify an error, highlight the relevant span of text, and select a category/sub-category and severity level from the available options. (The span of text may be in the source segment if the error is a source error or an omission.) When identifying errors, please be as fine-grained as possible. For example, if a sentence contains two words that are each mistranslated, two separate mistranslation errors should be recorded. If a single stretch of text contains multiple errors, you only need to indicate the one that is most severe. If all have the same severity, choose the first matching category listed in the error typology (eg, {\em Accuracy}, then {\em Fluency}, then {\em Terminology}, etc).\\
 
Please pay particular attention to document context when annotating. If a translation might be questionable on its own but is fine in the context of the document, it should not be considered erroneous; conversely, if a translation might be acceptable in some context, but not within the current document, it should be marked as wrong.\\
 
There are two special error categories: {\em Source error} and {\em Non-translation}. Source errors should be annotated separately, highlighting the relevant span in the source segment. They do not count against the 5-error limit for target errors, which should be handled in the usual way, whether or not they resulted from a source error. There can be at most one {\em Non-translation} error per segment, and it should span the entire segment. No other errors should be identified if {\em Non-Translation} is selected.
}}}
\caption{MQM annotator guidelines}
\label{tab:mqm-guidelines}
\end{table*}

Table~\ref{tab:mqm-severity} shows descriptions for three severity levels that raters must assign to errors independent of their category. Many MQM schemes include an additional “Critical” severity which is worse than Major, but we dropped this because its definition is often context-specific, capturing errors that are disproportionately harmful for a particular application. We felt that for broad coverage MT the distinction between Major and Critical was likely to be highly subjective, while Major errors (actual errors) would be easier to distinguish from Minor ones (imperfections). Neutral severity allows annotators to express subjective opinions about the translation without affecting its rating.

Annotator instructions are shown in Table~\ref{tab:mqm-guidelines}. We kept these minimal because our raters were professionals with previous experience in assessing translation quality, including with MQM. There are many subtle issues that arise in error annotation, such as the correct way to translate units (eg, should 1 inch be translated as 1 Zoll, 1cm, or 2.54cm?), but we resisted the temptation to establish an extensive list of context-specific guidelines, relying instead on the judgment of our annotators. In order to temper the effect of long segments, we imposed a maximum of five errors per segment. For segments with more errors, we asked raters to identify only the five most severe. Thus we do not distinguish between segments containing five or more than five Major errors, although we do distinguish between segments with many identifiable errors and those that are categorized as entirely Non-translation.
To focus our raters on careful error identification, and to provide potentially useful information for further studies, we had them highlight error spans in the text, following the conventions laid out in Table~\ref{tab:mqm-guidelines}.

\subsubsection*{Scoring}

Since we are ultimately interested in deriving scores for sentences, we require a weighting on error categories and severities. We set the weight on Minor errors to 1, and explored a range of Major error weights from 1 to 10 (the Major weight recommended in the MQM standard). For each weight combination we examined the stability of system ranking using a resampling technique. We found that a Major weight of 5 gave the best balance of stability and ability to discriminate among systems.

These weights apply to all error categories except Fluency/Punctuation and Non-translation. We assigned a weight of 0.1 for Fluency/Punctuation to reflect its mostly non-linguistic character. Decisions like the kind of quotation mark to use or the spacing between words and punctuation affect the appearance of a text but do not change its meaning. Unlike other kinds of minor errors, these are easy to correct algorithmically, so we assign them a low weight to ensure that their main role is to distinguish between systems that are equivalent in other respects. 
Our decision is supported by evidence from professional translators, who tend to treat minor punctuation errors as insignificant for the purpose of scoring, even when they are required to annotate them within the MQM framework. 
Note that this category does not include punctuation errors that render a text ungrammatical or change its meaning (eg, eliding the comma in “Let’s eat, grandma”), which have the same weight as other Major errors. Source errors are ignored in our current study but give us the ability to discard badly garbled source sentences, which might be prevalent in certain genres. The singleton Non-translation category has a weight of 25, equivalent to five Major errors, the worst segment-level score possible in our annotation scheme.

Our current weighting ignores the text span of errors, as this provides little information relevant to scoring once severity and category are taken into account.
 
Table~\ref{tab:mqm-scoring} summarizes our weighting scheme. The score of a segment is the sum of all errors it contains, averaged over all annotators, and ranges from 0 (perfect) to 25 (maximally bad). Segment scores are averaged to provide document- and system-level scores.

\section{Analysis of Metric Performance}
\label{sec:full-metrics}

\begin{figure*}[!ht]
    \centering
    \includegraphics[scale=0.4]{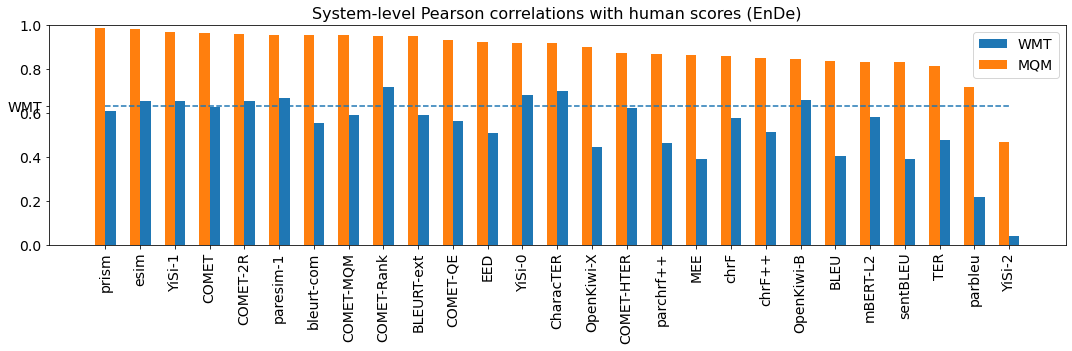}
    \includegraphics[scale=0.4]{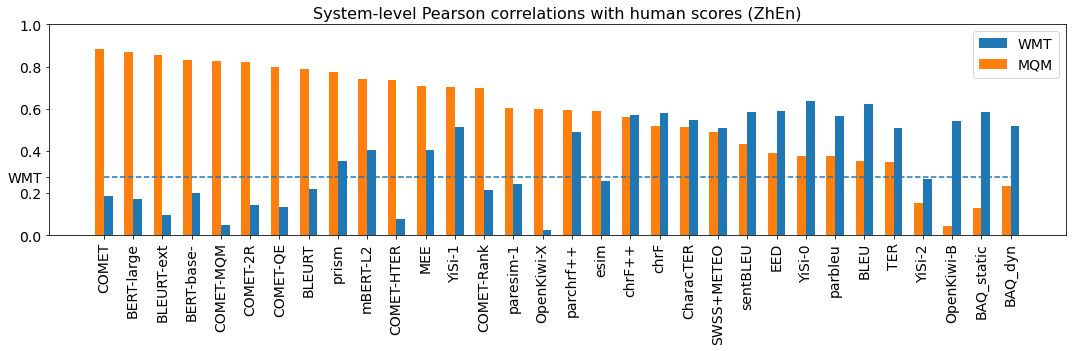}
    \caption{System-level Pearson correlation with MQM and WMT scoring.}
    \label{fig:metrics-sys-pearson}
\end{figure*}

\begin{figure*}[ht]
    \centering
    \includegraphics[scale=0.4]{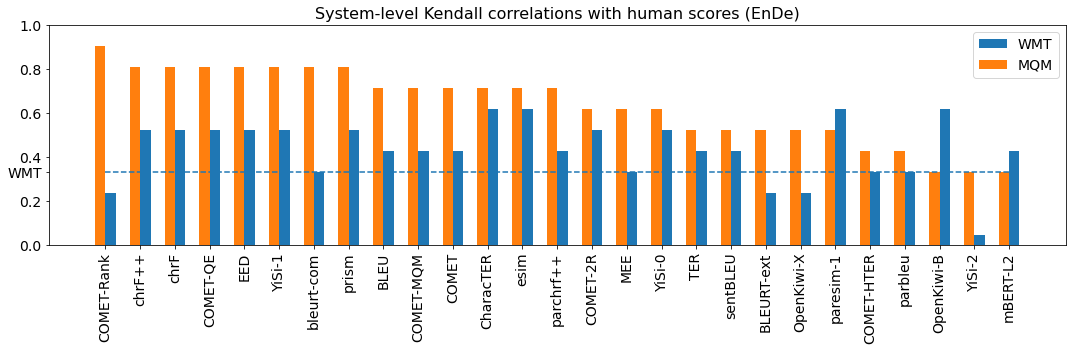}
    \includegraphics[scale=0.4]{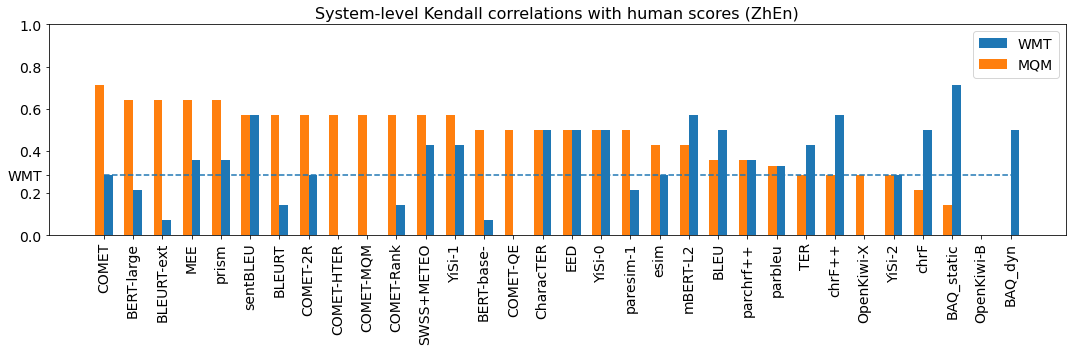}
    \caption{System-level Kendall correlation with MQM and WMT scoring.}
    \label{fig:metrics-sys-kendall}
\end{figure*}

\begin{figure*}[ht]
    \centering
    \includegraphics[scale=0.4]{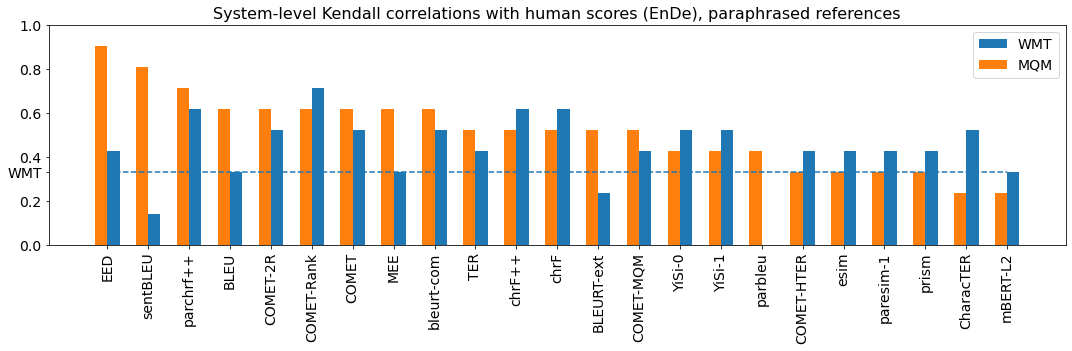}
    \caption{System-level Kendall correlation with MQM and WMT scoring when metrics use paraphrased reference.}
    \label{fig:metrics-sys-kendall-para}
\end{figure*}

\begin{figure*}[ht]
    \centering
    \includegraphics[scale=0.4]{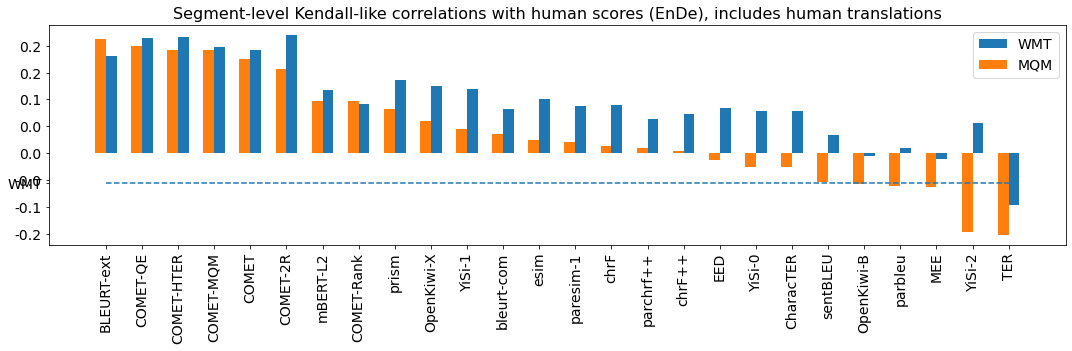}
    \includegraphics[scale=0.4]{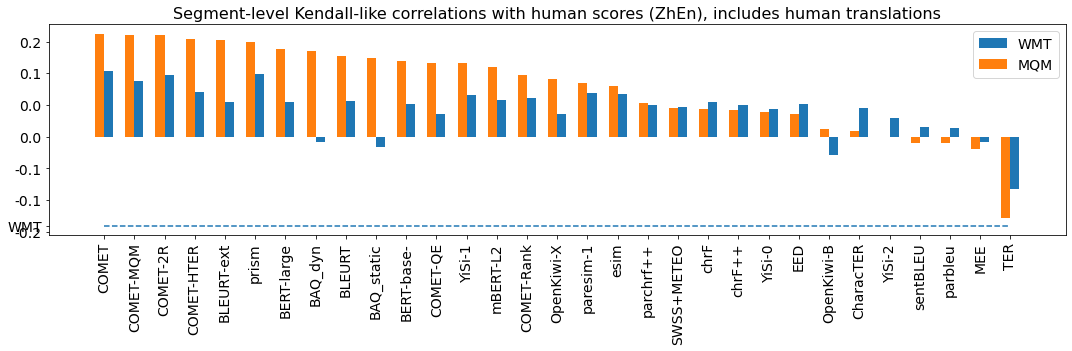}
    \caption{Segment-level Kendall correlation with MQM and WMT scoring when human outputs are included among systems to be scored.}
    \label{fig:metrics-seg-kendall-para-human}
\end{figure*}

Figure~\ref{fig:metrics-sys-kendall} shows the system-level Kendall tau correlations for all metrics from the WMT 2020 metrics task, completing the partial picture given in Figure~\ref{fig:metrics}. Figure~\ref{fig:metrics-sys-pearson} contains the corresponding plots for Pearson correlation. Figure~\ref{fig:metrics-sys-kendall-para} shows Kendall correlation for English$\to$German for metrics using the paraphrased references available for that language pair; this substantially changes metric ranking and performance. Finally, Figure~\ref{fig:metrics-seg-kendall-para-human} shows performance when human outputs were included among the systems to be scored, resulting in lower correlations compared to MQM gold scores, and much lower correlations compared to WMT gold scores.

\begin{figure*}[ht]
    \centering
    \includegraphics[scale=0.4]{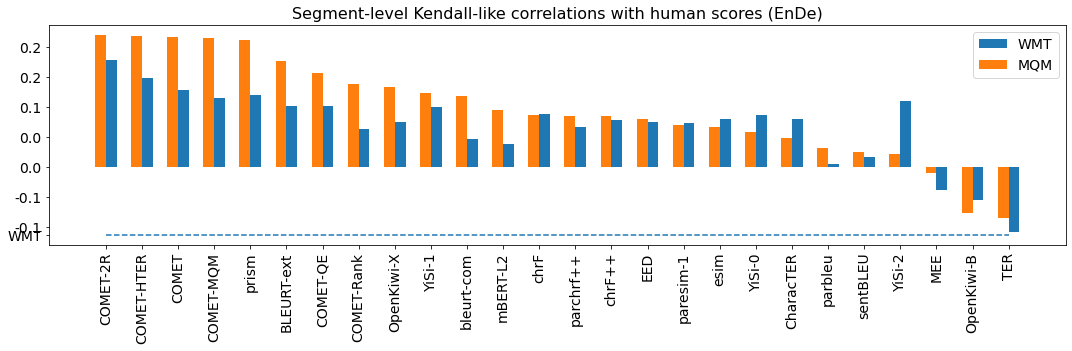}
    \includegraphics[scale=0.4]{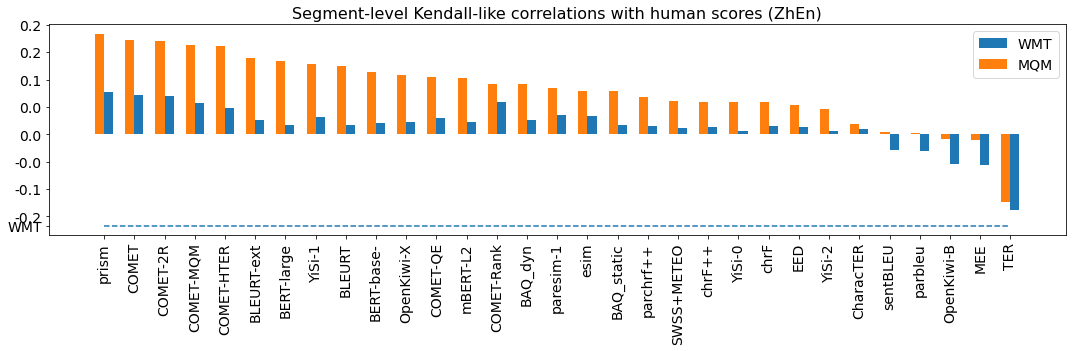}
    \caption{Segment-level Kendall correlation with MQM and WMT scoring.}
    \label{fig:metrics-seg-kendall}
\end{figure*}

\begin{figure*}[ht]
    \centering
    \includegraphics[scale=0.4]{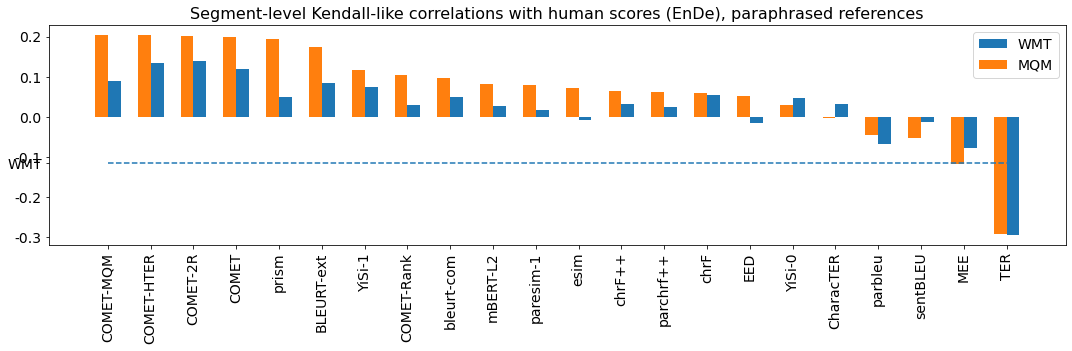}
    \caption{Segment-level Kendall correlation with MQM and WMT scoring when metrics use paraphrased reference.}
    \label{fig:metrics-seg-kendall-para}
\end{figure*}

For segment-level correlations, we adopted the WMT ``Kendall-like'' measure to deal with missing and unreliable segment-level annotations in the WMT data. This discards pairwise rankings when annotations are missing or when raw scores differ by less than 25. This statistic aggregates pairwise rankings over system scores for each segment rather than working from a single global list of segment-level scores, independent of which system they pertain to. For MQM correlations, lacking a way to establish a comparable threshold, and because we expected small differences to be significant, we used a threshold of 0. The results are shown in Figures~\ref{fig:metrics-seg-kendall}, \ref{fig:metrics-seg-kendall-para}, and \ref{fig:metrics-seg-kendall-para-human} for standard references, paraphrased references, and with human outputs included, respectively. In general, segment-level correlations are much lower than system-level, but patterns of differences between WMT and MQM correlations remain similar.

\begin{figure*}[ht]
    \centering
    \includegraphics[scale=0.4]{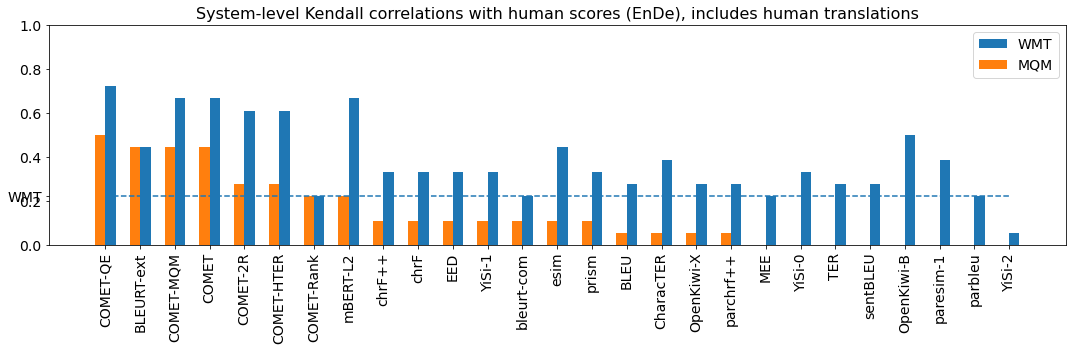}
    \includegraphics[scale=0.4]{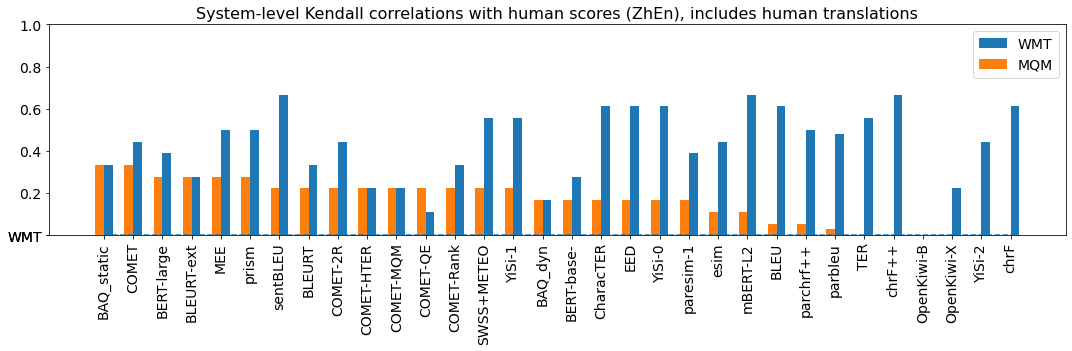}
    \caption{System-level Kendall correlation with MQM and WMT scoring when human outputs are included among systems to be scored.}
    \label{fig:metrics-sys-kendall-para-human}
\end{figure*}

\clearpage

\end{document}